\documentclass{article} 
\usepackage{iclr2027_conference,times}

\usepackage[utf8]{inputenc} 
\usepackage[T1]{fontenc}    
\usepackage{hyperref}       
\usepackage{url}            
\usepackage{booktabs}       
\usepackage{amsfonts}       
\usepackage{nicefrac}       
\usepackage{microtype}      
\usepackage{xcolor}         

\usepackage{graphicx}
\usepackage{subcaption}
\usepackage{amsmath}
\usepackage{bbm}
\usepackage{dsfont}
\usepackage{amssymb}
\usepackage{mathtools}
\usepackage{amsthm}
\usepackage{xspace}
\usepackage[capitalize,noabbrev]{cleveref}
\usepackage[textsize=tiny]{todonotes}

\theoremstyle{plain}
\newtheorem{theorem}{Theorem}[section]
\newtheorem{prop}[theorem]{Proposition}

\newtheorem{cor}[theorem]{Corollary}
\theoremstyle{definition}

\theoremstyle{remark}

\newtheorem{problem}[theorem]{Problem}

\newcommand{\nop}[1]{}

\newcommand{\ind}[1]{\mathds{1}_{\left[ {#1} \right]}}
\newcommand{\prob}{\mathbb{P}}

\title{Pay for Hints, Not Answers: LLM Shepherding for Cost-Efficient Inference}

\author{Ziming Dong$^{1}$, Hardik Sharma$^{2}$, Evan O'Toole$^{1}$, Jaya Prakash Champati$^{1}$, Kui Wu$^{1}$ \\
$^{1}$Department of Computer Science, University of Victoria, Victoria, Canada \\
$^{2}$Department of Information Technology, Manipal University, Manipal, India \\
\texttt{\{dzm1098,evanotoole,jpchampati,wkui\}@uvic.ca} \\
\texttt{hardik.23fe10ite00064@muj.manipal.edu}}

\iclrfinalcopy
\begin{document}

\maketitle
\lhead{}
\renewcommand{\headrulewidth}{0pt}

\begin{abstract}
Large Language Models (LLMs) deliver state-of-the-art performance on complex reasoning tasks, but their inference costs limit deployment at scale. Small Language Models (SLMs) offer dramatic cost savings yet lag substantially in accuracy. Existing approaches---routing and cascading---treat the LLM as an all-or-nothing resource: either the query bypasses the LLM entirely, or the LLM generates a complete response at full cost. We find that the LLM need not be used this way: even a truncated prefix of its response raises SLM accuracy, with gains that broadly grow with prefix length on all four benchmarks we study, also it hold across model families, tokenizers, and SLM sizes. We therefore introduce LLM Shepherding, which requests only a short prefix (a \emph{hint}) from the LLM and lets the SLM produce its own complete response. Shepherding generalizes both routing and cascading, and it achieves lower cost under oracle decision-making. We develop a two-stage predictor that jointly determines whether a hint is needed and how many tokens to request. On GSM8K, CNK12, HumanEval, and MBPP, we compare shepherding with routing and cascading. Across full cost--accuracy sweeps, Reactive Shepherding attains the largest area under the cost--accuracy curve on all four benchmarks; at a fixed target of 90\% of LLM accuracy, it is the cheapest method on all four; and at default operating points it reduces cost by 33--65\% relative to LLM-only inference and delivers up to 4$\times$ cost reduction compared to state-of-the-art routing and cascading baselines. To our knowledge, shepherding is the first to formalize and learn a single hand-off prefix hint, controlled only through \texttt{max\_new\_tokens}, under a black-box per-token cost model.
\end{abstract}


\section{Introduction}

LLMs excel on complex reasoning tasks, but their inference costs limit deployment at scale. Meanwhile, the emergence of highly capable open-source Small Language Models (SLMs) has created new opportunities for cost-efficient inference on edge devices, private servers, or shared datacenters. SLMs offer clear advantages: lower latency, improved privacy, and significantly smaller monetary and energy costs. However, for tasks involving logic and mathematics, the quality gap between SLMs and frontier LLMs (e.g., GPT, Gemini, Claude) remains substantial~\citep{subramanian2025small}. Users and organizations thus face a difficult trade-off: accept inferior response quality to reduce costs, or pay a premium for LLM access to maintain accuracy.

This cost-quality trade-off motivates a central question: \emph{How can we substantially improve SLM output quality while minimizing reliance on expensive LLMs?} The question is especially pressing in resource-constrained settings—edge devices with limited connectivity, latency-sensitive applications, and cost-conscious deployments—where every LLM call carries significant overhead. The goal is not merely faster inference, but fundamentally less LLM computation for a given quality target.

\begin{figure*}[t]
    \centering
    \begin{subfigure}[b]{0.49\textwidth}
        \centering
        \includegraphics[width=\textwidth]{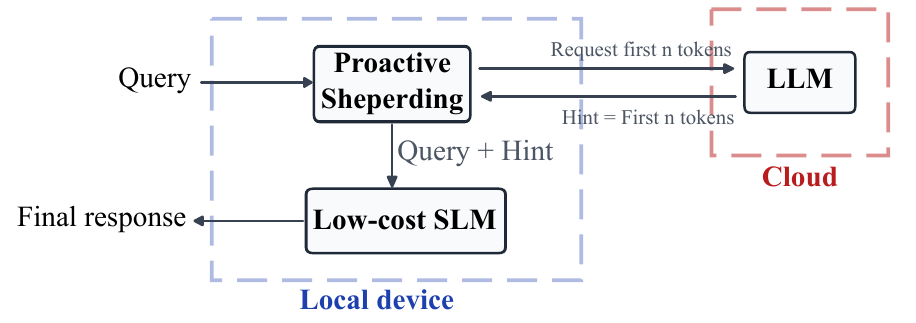}
        \caption{Proactive Shepherding}
        \label{fig:proactive-shepherding}
    \end{subfigure}
    \hfill
    \begin{subfigure}[b]{0.49\textwidth}
        \centering
        \includegraphics[width=\textwidth]{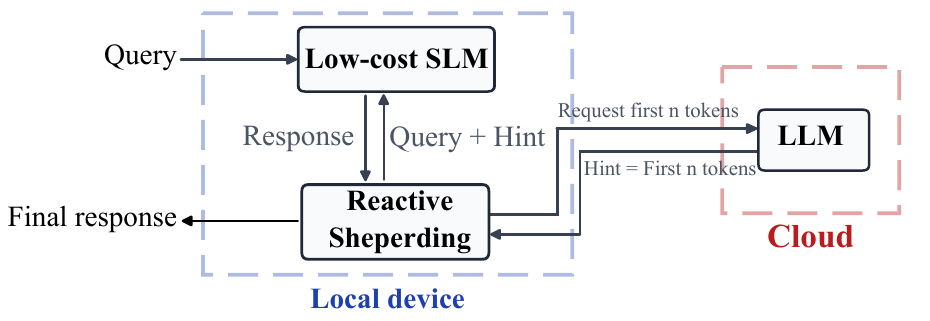}
        \caption{Reactive Shepherding}
        \label{fig:reactive-shepherding}
    \end{subfigure}
    \caption{Two modes of LLM Shepherding. Both request the first $n$ LLM tokens as a hint for the SLM. (a) Proactive: decides upfront. (b) Reactive: decides after SLM failure.}
    \label{fig:shepherding-approaches} 
\end{figure*}

Two complementary paradigms have emerged to address this challenge: routing and cascading~\citep{behera2025towards}. \emph{Routing}~\citep{lu2023routing,ding2024hybrid,ong2024routellm} uses a learned classifier to direct each query to exactly one model—either the SLM handles the query entirely, or it is forwarded to the LLM based on query complexity. \emph{Cascading}~\citep{chen2024frugalgpt,aggarwal2024automix,gupta2024language} takes a sequential approach: the SLM attempts a response first, and the LLM is invoked only if the SLM response fails a confidence or correctness check.

We observe that both routing and cascading share a fundamental limitation: they treat the LLM as an all-or-nothing resource. Either the query bypasses the LLM entirely, or the LLM generates a complete response, incurring the full LLM inference cost. This binary view overlooks a key opportunity: \emph{most LLM service providers allow users to specify \texttt{max\_new\_tokens}, a user-defined limit for the number of output tokens}. This capability can be strategically leveraged to reduce LLM costs while simultaneously improving SLM answer quality.

\begin{figure}
    \centering
    \includegraphics[width=0.6\linewidth]{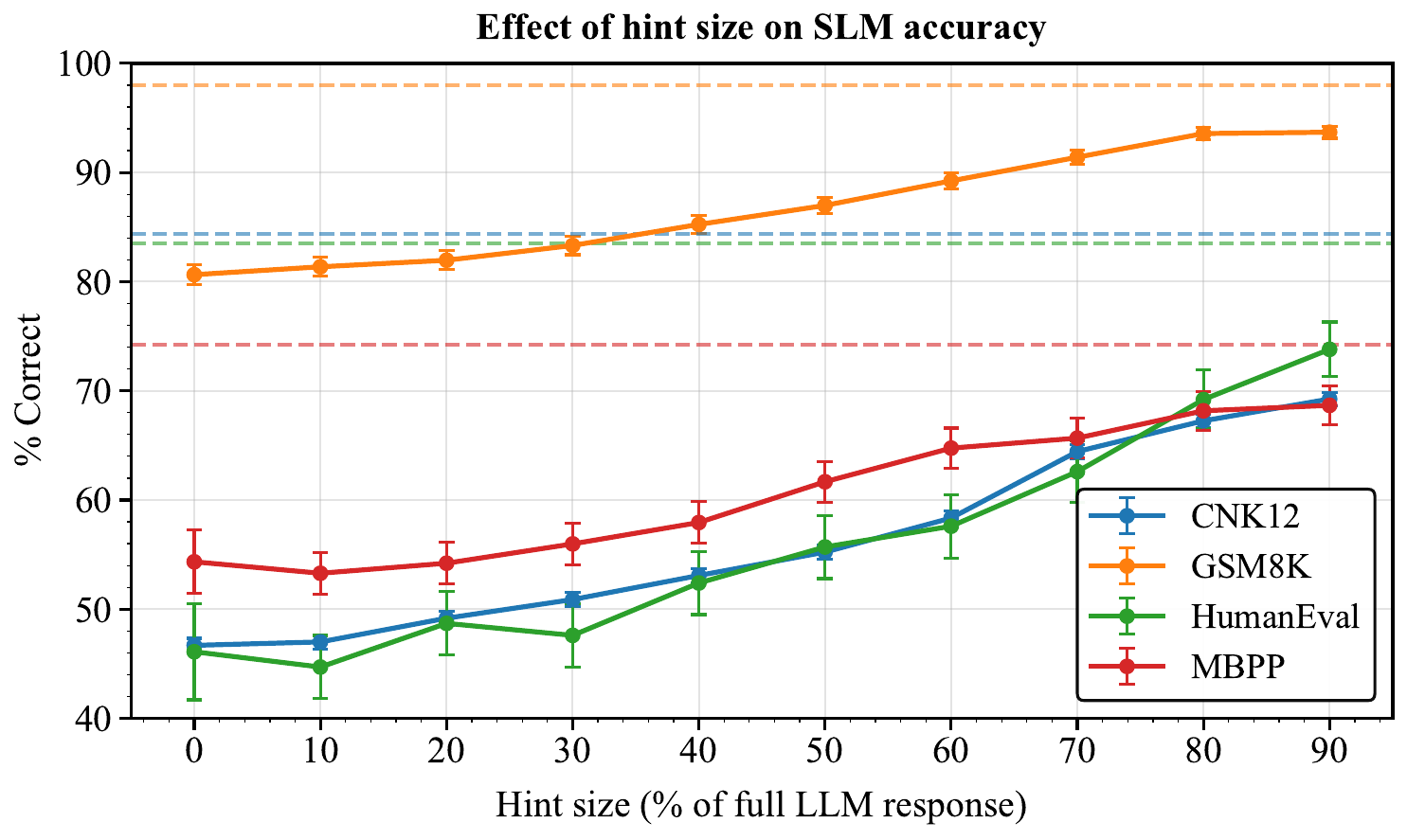}
    \caption{SLM (Llama-3.2-3B-Instruct) accuracy vs.\ LLM (Llama-3.3-70B-Versatile) hint size. Accuracy broadly rises with hint size, and on the math benchmarks the gains begin with the smallest hints. Dashed lines indicate LLM accuracy.} \label{fig:hints-help}
    \vspace{-0.15in}
\end{figure}

We propose \textbf{LLM Shepherding}, a novel framework that complements routing and cascading by enabling \emph{partial} LLM assistance. Rather than having the LLM produce complete responses, shepherding requests the LLM to output only a short prefix—first $n$ tokens of the LLM response—called a \emph{hint}. The hint is then concatenated with the query and is given to SLM for the final response. To validate whether LLM hints can benefit SLM inference, we evaluate on four widely-used benchmarks spanning mathematical reasoning—GSM8K~\citep{cobbe2021gsm8k} and CNK12~\citep{openr1_math_220k}—and code generation—HumanEval~\citep{chen2021evaluatinglargelanguagemodels} and MBPP~\citep{austin2021programsynthesislargelanguage}. We use Llama-3.2-3B-Instruct as the SLM and Llama-3.3-70B-Versatile as the LLM~\citep{dubey2024llama}. As shown in Figure~\ref{fig:hints-help}, SLM accuracy rises with hint size on all four benchmarks. The rise is steady on the math benchmarks, whereas on the code benchmarks hints below roughly 40\% of the response give no reliable gain and larger hints give large ones. We further verify in Appendix~\ref{appendix:cross-model} that this broadly increasing, though not strictly monotonic, hint-size effect holds for cross-family SLM-LLM pairs\nop{ (e.g., Llama SLM with Gemma LLM)}, which suggests it is a property of prefix-guided SLM generation rather than of our specific setup.

As illustrated in Figure~\ref{fig:shepherding-approaches}, we instantiate two modes of shepherding: 1) \textbf{Proactive Shepherding} (Figure~\ref{fig:proactive-shepherding}) is a \emph{routing-based} shepherding, where a learned router directly predicts whether a hint is needed and how many tokens to request, and 2) \textbf{Reactive Shepherding} (Figure~\ref{fig:reactive-shepherding}) is a \emph{cascading-based} shepherding, where the features of the SLM response(s) are used for cascading decision and hint prediction.
Observe that proactive shepherding reduces to routing at the extremes: hint size zero (SLM only) or maximal (full LLM response). Similarly, reactive shepherding reduces to cascading when a complete LLM output is requested upon SLM failure. Shepherding thus generalizes both paradigms. By operating in the intermediate regime -- requesting hints rather than full responses -- shepherding theoretically achieves lower costs than either approach.


Realizing this vision in practice presents substantial challenges. First, the minimum hint size is zero-inflated across queries: on GSM8K, 80\% of queries need no hint, while the conditional distribution over positive hint sizes spans 10--90\% of the full LLM response. Second, the benefit of additional tokens is non-smooth—critical information may appear at specific token boundaries, causing abrupt quality transitions. Third, SLM accuracy does not increase steadily with hint length: on HumanEval it fluctuates within the error bars at small and intermediate budgets and peaks only at the largest one (Figure~\ref{fig:hints-help}), so the minimum sufficient size cannot be read off a smooth curve.

These challenges motivate our two-stage prediction framework that jointly determines whether a hint is needed and how many tokens to request.

\textbf{Contributions:}
\begin{itemize}
    \item We formalize LLM Shepherding and prove that, under oracle decision-making and a mild condition on token prices, it costs no more than routing and cascading (Section~\ref{sec:problem_formulation}).
    
    \item We develop a two-stage prediction model that jointly predicts whether a query requires a hint and how many tokens to request, addressing key challenges including a non-smooth accuracy response and zero-inflated hint size distributions(Section~\ref{sec:training}).
    
    \item We empirically evaluate shepherding against state-of-the-art routing strategies (RouteLLM~\citep{ong2024routellm}, GraphRouter~\citep{GraphRouter}) and cascading strategies (FrugalGPT~\citep{chen2024frugalgpt}, ABC~\citep{kolawole2025agreementbased}) on mathematical reasoning (GSM8K, CNK12) and code generation (HumanEval, MBPP) benchmarks. Reactive Shepherding attains the highest AIQ on all four benchmarks, is the cheapest method at a 90\% accuracy target on all four, and achieves the highest ACE at default operating points, with four times the cost reduction of routing baselines on the CNK12 and 1.5 times that of cascading baselines on HumanEval, the latter without any code-specific training (Section~\ref{sec:performance}).
\end{itemize}

\section{Problem Formulation}
\label{sec:problem_formulation}


\subsection{System and Notation}

Let $\mathcal{Q}$ denote the set of all possible queries and let $q \in \mathcal{Q}$ denote an individual query. Let $\mathcal{X}$ denote a finite vocabulary of tokens and $\mathcal{X}^*$ denote the set of all finite token sequences over $\mathcal{X}$. We consider two language models: an open-source SLM $h_s: \mathcal{Q} \rightarrow \mathcal{X}^*$ deployed on a local device, such as a smartphone, laptop, or local server, and an LLM $h_l: \mathcal{Q} \rightarrow \mathcal{X}^*$ accessible through an LLM API.

For a token sequence $x \in \mathcal{X}^*$, we use $|x|$ to denote its number of tokens. For two sequences $x, y \in \mathcal{X}^*$, we denote their concatenation as $x \oplus y$. Given a sequence $x$ and an integer $n \leq |x|$, $x_{[1:n]}$ denotes the prefix of $x$ consisting of its first $n$ tokens. We use $\mathds{1}[\cdot]$ to denote the indicator function.


We define a quality function $\phi: \mathcal{Q} \times \mathcal{X}^* \rightarrow [0, 1]$ that measures the quality of a response $r \in \mathcal{X}^*$ to a query $q \in \mathcal{Q}$. Higher values indicate better response quality. In practice, $\phi$ may be instantiated as an exact-match accuracy metric (for tasks with ground-truth answers), a semantic similarity score, or a learned reward model.
\nop{We define a quality threshold $\tau \in (0, 1]$ such that a response $r$ is deemed \emph{satisfactory} for query $q$ if $\phi(q, r) \geq \tau$.

We introduce the indicator function $I_s: \mathcal{Q} \rightarrow \{0, 1\}$, where:
\begin{equation}
I_s(q) = \begin{cases}
1 & \text{if } \phi(q, h_s(q)) \geq \tau \\
0 & \text{otherwise}
\end{cases}
\end{equation}
That is, $I_s(q) = 1$ if the SLM produces a satisfactory response to query $q$ without assistance, and $I_s(q) = 0$ otherwise.
}
We define a quality threshold $\tau \in (0, 1]$ and $I_s(q) = \ind{\phi(q, h_s(q)) \geq \tau}$ to indicate whether the SLM alone produces a satisfactory response.

\subsection{Hints and Shepherding Policy}

The central concept in LLM shepherding is the \emph{hint}, defined as a prefix of the LLM's response provided to the SLM to guide its generation. Formally, for a query $q$ and hint size $n \geq 0$, the hint is given by $\text{hint}(q, n) = h_l(q){[1:n]}$.
The hint-augmented SLM response is then:
\begin{equation}\label{eq:oplus}
h_s^{(n)}(q) = h_s\bigl(q \oplus \text{hint}(q, n)\bigr), 
\end{equation}
where the SLM generates a response for the original query concatenated with the hint prefix.


\paragraph{Shepherding Policy.}
A \emph{shepherding policy} $\pi: \mathcal{Q} \to \{0, 1, \ldots, N_{\max}\}$ maps query $q$ to a hint size $n = \pi(q)$, where $N_{\max}$ is the maximum output token limit of the LLM. When $n = 0$, the SLM responds without assistance; when $n > 0$, the system requests an $n$-token hint before SLM completion. For each query $q$, we define the \emph{minimum hint size} as the smallest number of LLM tokens required for the SLM to produce a satisfactory response:
\begin{equation}\label{eq:n*}
n^*(q) = \min \left\{ n \in \mathbb{Z}_{\geq 0} : \phi\bigl(q, h_s^{(n)}(q)\bigr) \geq \tau \right\}.
\end{equation}
The \textit{Oracle Shepherding Policy} $\pi^*$ maps each query $q$ to $n^*(q)$.

\subsection{Cost Model and Evaluation Metrics}

We adopt a token-based cost model consistent with commercial pricing. Let $c_l^{\text{in}}$ and $c_l^{\text{out}}$ denote the cost per input token and output token for the LLM, respectively. Similarly, let $c_s^{\text{in}}$ and $c_s^{\text{out}}$ denote the corresponding costs for the SLM. Both LLM and SLM costs are treated as general non-negative quantities, capturing deployments where the SLM runs on metered local hardware, private infrastructure, or shared datacenters. Let $c_\pi(q)$ denote the cost that a policy $\pi$ incurs on query $q$, and let $C_\pi=\sum_{q\in\mathcal{Q}} c_\pi(q)$ denote its total cost over $\mathcal{Q}$. For a shepherding policy, the explicit form of $c_\pi(q)$ differs between proactive and reactive shepherding, and Proposition~\ref{prop:oracle_costs} in Section~\ref{sec:costanalysis} gives both forms under oracle decision-making. As reference points, the LLM-only and SLM-only strategies incur per-query costs $c_l(q)=|q|\,c^{\text{in}}_l+|h_l(q)|\,c^{\text{out}}_l$ and $c_s(q)=|q|\,c^{\text{in}}_s+|h_s(q)|\,c^{\text{out}}_s$, with total costs $C_l$ and $C_s$ obtained by summing over $\mathcal{Q}$.

\paragraph{Evaluation metrics.}
Let $A_l$ and $A_s$ denote the accuracies of the LLM and the SLM, and $A_\pi$ the accuracy of a policy $\pi$. Every strategy we compare exposes hyperparameters that trade cost against accuracy, so we evaluate strategies at three levels, in the order used in Section~\ref{sec:performance}: across the full operating range, at a fixed accuracy target, and at a single default operating point.

\nop{\emph{Full operating range.} Sweeping the hyperparameters of a strategy yields a set of policies $\Pi$, and each $\pi\in\Pi$ gives one operating point $(C_\pi, A_\pi)$. These points give the strategy's cost--accuracy curve $A_\Pi(c)=\max\bigl(\{A_s\}\cup\{A_\pi : \pi\in\Pi,\ C_\pi\le c\}\bigr)$, the highest accuracy the strategy attains within a total budget $c$, over the cost interval $[0, C_l]$ shared by all strategies (Section~\ref{sec:pareto}). A common interval makes values directly comparable across strategies, and crediting only accuracies reached by a single configuration, with the SLM-only accuracy as a floor, reflects how strategies are deployed in practice. We summarize each curve by its Average Improvement in Quality (AIQ)~\citep{hu2024routerbench}, the area under the curve normalized by the length of the interval:
\begin{equation}\label{eq:AIQ}
\mathrm{AIQ}(\Pi) = \frac{1}{C_l}\int_{0}^{C_l} A_\Pi(c)\,dc .
\end{equation}
}

\emph{Full operating range.} Sweeping a strategy's hyperparameters yields a set of policies $\Pi$, each giving an operating point $(C_\pi, A_\pi)$. The strategy's cost--accuracy curve $A_\Pi(c)=\max\bigl(\{A_s\}\cup\{A_\pi : \pi\in\Pi,\ C_\pi\le c\}\bigr)$ is the highest accuracy it attains within a total budget $c$, with the SLM-only accuracy as a floor (Section~\ref{sec:pareto}). We summarize each curve by its Average Improvement in Quality (AIQ)~\citep{hu2024routerbench}, the area under the curve normalized over the interval $[C_s, C_l]$ shared by all strategies:
\begin{equation}\label{eq:AIQ}
\mathrm{AIQ}(\Pi) = \frac{1}{C_l - C_s}\int_{C_s}^{C_l} A_\Pi(c)\,dc .
\end{equation}
The interval starts at $C_s$, the cheapest budget at which the SLM-only policy, and hence the floor $A_s$, is attainable.

\nop{\emph{Fixed accuracy target.} Many deployments must meet a fixed quality level, such as a service-level agreement. For a query distribution $\mathcal{D}$ and a target expected quality $\bar\tau$, we therefore also compare the minimum cost each strategy needs to reach that target (Section~\ref{min}):
\begin{equation}\label{eq:cost_optimal}
\min_{\pi}\ \mathbb{E}\big[c_\pi(q)\big] \quad \text{s.t.}\quad \mathbb{E}\big[\phi\big(q,h^{(\pi(q))}_s(q)\big)\big]\ \ge\ \bar\tau .
\end{equation}
}

\emph{Fixed accuracy target.} Many deployments must meet a fixed quality level, such as a service-level agreement. For a target accuracy $\bar\tau$, we therefore also compare the minimum cost each strategy needs to reach it (Section~\ref{min}):
\begin{equation}\label{eq:cost_optimal}
C^*_\Pi(\bar\tau) = \min_{\pi\in\Pi}\ C_\pi \quad \text{s.t.}\quad A_\pi \ge \bar\tau ,
\end{equation}
which is undefined when no configuration of the strategy reaches $\bar\tau$.

\emph{Single operating point.} Finally, to compare strategies at their default settings (Section~\ref{sec:default-points}), we use Accuracy-per-Cost Efficiency (ACE):
\begin{equation*}
\mathrm{ACE}(\pi) = \frac{(A_\pi - A_s)/(A_l - A_s)}{(C_\pi - C_s)/(C_l - C_s)},
\end{equation*}
which measures the fraction of the SLM--LLM accuracy gap a policy recovers per unit of the SLM--LLM cost gap it spends. ACE coincides with the IBC metric of AutoMix~\citep{aggarwal2024automix} up to a fixed normalization; Appendix~\ref{app:ace} discusses its relation to IBC and PGR/APGR~\citep{ong2024routellm}. ACE is undefined for the SLM-only policy, whose numerator and denominator both vanish; it is the reference point rather than a competitor.

\nop{
\subsection{\color{blue} Challenges \color{black}}

Solving the above optimization problems exactly is intractable for several reasons:

\begin{enumerate}
    \item \textbf{Query-dependent difficulty}: The minimum sufficient hint size $n^*(q)$ varies significantly across queries. Simple queries may require no hint ($n^*(q) = 0$), while complex reasoning tasks may require substantial guidance.

    \item \textbf{Non-monotonic quality}: The quality function $\phi(q, h_s^{(n)}(q))$ is not necessarily monotonically increasing in $n$. Providing too many hint tokens may constrain the SLM's generation in unintended ways.

    \item \textbf{Unknown quality function}: The quality function $\phi$ is typically not available at inference time, necessitating a learned predictor.
\end{enumerate}

These challenges motivate our approach of learning a hint router that jointly predicts (i) whether a query requires a hint, and (ii) if so, how many hint tokens to request. We describe this approach in detail in the next section.
}

\subsection{Cost Analysis (Oracle)} \label{sec:costanalysis}
We analyze the monetary cost of routing, cascading, and the two modes of shepherding Proactive and Reactive which under oracle decision-making. We assume the oracle has perfect knowledge of $I_s(q)$ and $n^*(q)$. Routing incurs either the full SLM cost or the full LLM cost (otherwise), but never both. Cascading always incurs the SLM cost first; if the SLM fails, it additionally incurs the full LLM cost.

Proactive Shepherding makes the hint decision upfront: when $n^*(q)=0$ only the SLM is invoked; otherwise the system pays (i) the LLM input cost for processing the query, (ii) the LLM output cost for generating $n^*(q)$ hint tokens, and (iii) the SLM cost on the query augmented with the hint.
Reactive Shepherding follows the cascading structure: it first invokes the SLM, and only when $I_s(q)=0$ does it additionally pay the LLM input/hint-output cost and re-run the SLM on the augmented input.
This yields the following proposition.

\begin{prop}[Oracle Costs]
\label{prop:oracle_costs}
Under oracle decision-making, the monetary cost per query $q \in \mathcal{Q}$ for routing, cascading, Proactive Shepherding, and Reactive Shepherding is:
\begin{align*}
&c_{\text{route}}^*\!(q) \!=\! I_s(q)\,(|q| c_s^{\text{in}}\! +\! |h_s(q)| c_s^{\text{out}}) \!+\! \big(1 \!-\! I_s(q)\big)(|q| c_l^{\text{in}} \!+\! |h_l(q)| c_l^{\text{out}}) \\
&c_{\text{casc}}^*\!(q) \!=\! |q| c_s^{\text{in}} \!+\! |h_s(q)| c_s^{\text{out}} \!+\! \big(1 \!-\! I_s(q)\big)(|q| c_l^{\text{in}} \!+\! |h_l(q)| c_l^{\text{out}}) \\
&c_{\text{proac-shep}}^*\!(q) \!=\! \ind{n^*(q) > 0} |q| c_l^{\text{in}} \!+\! n^*\!(q) c_l^{\text{out}} \!+\! (|q| \!+\! n^*\!(q)) c_s^{\text{in}} \!+\! |h_s^{(n^*\!(q))}\!(q)| c_s^{\text{out}} \\
&c_{\text{reac-shep}}^*\!(q) \!=\! |q| c_s^{\text{in}} \!+\! |h_s(q)| c_s^{\text{out}} \!+\! \big(1 \!-\! I_s(q)\big) \big[|q| c_l^{\text{in}} \!+\! n^*\!(q) c_l^{\text{out}} +\! (|q| \!+\! n^*\!(q)) c_s^{\text{in}} \!+\! |h_s^{(n^*\!(q))}\!(q)| c_s^{\text{out}}\big],
\end{align*}
where $h_s^{(n)}(q)$ is as defined in \eqref{eq:oplus}.
\end{prop}

The following corollary identifies the condition under which each mode of shepherding dominates its corresponding baseline (proof in Appendix~\ref{proof:cor:dominance}).

\begin{cor}
\label{cor:dominance}
For every query $q \in \mathcal{Q}$ with $I_s(q) = 0$, both
\[
c_{\text{proac-shep}}^*(q) \leq c_{\text{route}}^*(q) \quad \text{and} \quad c_{\text{reac-shep}}^*(q) \leq c_{\text{casc}}^*(q)
\]
hold if and only if
\begin{equation}
\big(|h_l(q)| - n^*(q)\big)\, c_l^{\text{out}} \;\geq\; (|q| + n^*(q))\, c_s^{\text{in}} + |h_s^{(n^*(q))}(q)|\, c_s^{\text{out}}.
\label{eq:dominance_condition}
\end{equation}
For queries with $I_s(q) = 1$, all four costs reduce to the SLM-only cost $|q|\, c_s^{\text{in}} + |h_s(q)|\, c_s^{\text{out}}$ and the inequalities hold with equality.
\end{cor}

The condition in~\eqref{eq:dominance_condition} means that shepherding pays less than its baseline whenever the savings from generating fewer LLM output tokens, $(|h_l(q)| - n^*(q))\, c_l^{\text{out}}$, outweigh the additional SLM cost incurred from processing the longer (query + hint) input and producing an SLM response. Since LLM output tokens are typically the dominant cost factor (with $c_l^{\text{out}} > c_l^{\text{in}}$ and $c_l^{\text{out}} \gg c_s^{\text{in}}, c_s^{\text{out}}$ in practice), \eqref{eq:dominance_condition} is satisfied for the vast majority of queries.

\section{Training the Shepherding System}
\label{sec:training}

Our goal is to learn a policy $\pi_\theta:\mathcal{Q}\rightarrow \mathbb{Z}_{\ge 0}$ that predicts the minimum hint size $n^*(q)$ defined in Eq.~\eqref{eq:n*}. Direct prediction of $n^*(q)$ is challenging, because 1) \textbf{Zero-inflated hint sizes:} $n^*(q)$ has a large point mass at zero—on GSM8K, 80.6\% of queries require no hint at all—while the conditional distribution over positive hint sizes spans 10\% to 90\% of the full LLM response with approximately exponential decay (see Appendix~\ref{app:heavy_tail}); 2) \textbf{Non-smooth benefit:} a small change in $n$ can abruptly determine whether the SLM crosses the quality threshold $\tau$, as critical information may appear at specific token boundaries; 3) \textbf{Noisy accuracy curve:} measured accuracy does not increase steadily with $n$; on HumanEval it fluctuates within the error bars at small and intermediate budgets and peaks only at the largest one (cf.~Fig.~\ref{fig:hints-help}); and 4) \textbf{Tokenization sensitivity:} the same semantic content yields different token counts across models and query phrasings, making direct regression to absolute counts brittle. See Appendix~\ref{app:prediction_challenges} for more discussion.

We address them with a two-step policy: first predict whether a hint is needed via a binary indicator $y(q) = \ind{n^*(q) > 0}$, then regress the hint size on positive samples only. Note that $y(q) = 1 - I_s(q)$: a query requires a hint if and only if the SLM alone fails to produce a satisfactory response.\nop{, so queries with $y(q) = 0$ can be handled without incurring any LLM cost.}

\subsection{Training Data and Supervision Signals}

For each query $q$, we construct supervision labels by evaluating candidate hints at percentage sizes $p \in \{0, 10, 20, \ldots, 90\}$ of the full LLM response length $|h_l(q)|$, computing $n_p = \lfloor \frac{p}{100}|h_l(q)| \rfloor$ and querying the LLM with \texttt{max\_new\_tokens} $= n_p$ to obtain $\text{hint}(q, n_p) = h_l^{(n_p)}(q)$. We record
\[
n^*(q) = \min\left\{n_p : \phi\bigl(q, h_s(q \oplus \text{hint}(q, n_p))\bigr) \ge \tau\right\},
\]
falling back to $n^*(q) = |h_l(q)|$ if no partial hint meets the threshold. The full labeling procedure, outlier filtering (Appendix~\ref{Outlier}), discretization rationale, and decoding configuration are detailed in Appendix~\ref{app:training_dataset}. The log-transformed regression target is $r(q) = \log(1 + n^*(q))$.

\subsection{Model and Training Objective}
We use DeBERTa-v3-large \citep{he2023debertav3improvingdebertausing} as the text encoder and fuse the \texttt{[CLS]} embedding $\mathbf{h}_{\text{CLS}}(q)$ with transformed numeric features $\mathbf{g}(q)=\mathrm{MLP}(\mathbf{f}(q))$ into $\mathbf{u}(q)=[\mathbf{h}_{\text{CLS}}(q);\mathbf{g}(q)]$. Two lightweight heads produce the hint-indicator logit $\widehat{y}(q) = \mathrm{Head}_{\text{hint}}(\mathbf{u}(q))$ and the size estimate $\widehat{r}(q) = \mathrm{Head}_{\text{size}}(\mathbf{u}(q))$, with multi-sample dropout \citep{inoue2019multisample} on the indicator logit to reduce variance (details in Appendix~\ref{app:architecture}). The two heads are trained jointly with $\mathcal{L} = \lambda\,\mathcal{L}_{\text{hint}} + (1-\lambda)\,\mathcal{L}_{\text{size}}$. Here $\mathcal{L}_{\text{hint}}$ is the binary cross-entropy between $\sigma(\widehat{y}(q))$ and $y(q)$. $\mathcal{L}_{\text{size}}$ is the Smooth L1 (Huber) loss \citep{huber1964robust} between $\widehat{r}(q)$ and $r(q)$, averaged over the positive set $\mathcal{Q}^+ = \{q \mid y(q)=1\}$ because the hint size is undefined for no-hint queries. The weight $\lambda \in [0,1]$ is tuned on the validation set, and class imbalance is handled by weighted minibatch sampling. The AdamW/EMA setup and policy calibration are given in Appendix~\ref{opt}.

\subsection{Inference: Mapping Predictions to Decisions}

At test time, let $\prob(q) = \sigma(\widehat{y}(q))$ and apply threshold $\alpha \in [0,1]$:
\[
\pi_\theta(q) =
\begin{cases}
0, & \text{if } \prob(q) < \alpha,\\[4pt]
\left\lfloor \mathrm{clip}\!\left(\exp(\widehat{r}(q)) - 1,\, 0,\, N_{\max}\right)\right\rceil, & \text{if } \prob(q) \geq \alpha,
\end{cases}
\]
where $\lfloor x \rceil$ denotes rounding to the nearest integer, $\mathrm{clip}(x, a, b) = \max(a, \min(x, b))$ clamps values to $[a,b]$, and $N_{\max}$ is the maximum LLM output token limit. The parameter $\theta$ encompasses all learned components (backbone encoder, MLP, and prediction heads).

\noindent\textbf{Hint Threshold.} For queries with $\pi_\theta(q) = 0$, we return the SLM output $h_s(q)$ directly. For queries requiring hints ($\pi_\theta(q) > 0$), we further apply a hint-size threshold $\eta_{\text{hint}}$ to distinguish low- from high-complexity queries:
\begin{itemize}
    \item If $0 < \pi_\theta(q) \le \eta_{\text{hint}}$, we query the LLM with $\texttt{max\_new\_tokens} = \pi_\theta(q)$ to obtain $\text{hint}(q, \pi_\theta(q)) = h_l^{(\pi_\theta(q))}(q)$, and return the shepherded response $h_s(q \oplus \text{hint}(q, \pi_\theta(q)))$;
    \item If $\pi_\theta(q) > \eta_{\text{hint}}$, the query is deemed high-complexity and routed directly to the LLM, returning $h_l(q)$ without invoking the SLM.
\end{itemize}
This threshold-based routing means that queries requiring extensive hints may be better served by the LLM's full reasoning capacity, avoiding the overhead of generating a lengthy hint only to invoke the SLM. The threshold $\eta_{\text{hint}}$ is tuned on the validation set; see Appendix~\ref{app:inference} for further rationale. 

\subsection{Reactive Shepherding: Cascading Decision} \label{sec:reactive}

For reactive shepherding, we add a training-free cascading module before the shepherding model, following the agreement-based approach of \citep{kolawole2025agreementbased}. We run the SLM $K$ times ($K \in \{2,3\}$, listed per dataset in Table~\ref{tab:method-configs}) with temperature sampling ($T=0.3$) to generate candidate answers $\{A_1,\ldots,A_K\}$; if at least $k$ answers agree on the final answer, we return that answer directly, otherwise the query is passed to the shepherding model. We additionally incorporate features derived from the $K$ responses---average predictive entropy, average output length, and query length---into the shepherding model's numeric input $\mathbf{f}(q)$. Intuitively, higher entropy signals greater SLM uncertainty and thus a stronger need for a hint, while output-length statistics provide a proxy for the verbosity and reasoning depth required by the query. The number of samples $K$ and the agreement requirement $k$ trade precision against routing rate. The values of $K$ we use are listed in Appendix~\ref{app:method-configs}, and Appendix~\ref{app:pareto:setup} sweeps $k$. Full feature definitions and design justifications are in Appendix~\ref{app:reactive_features}.
\section{Performance Comparison} \label{sec:performance}

We compare the cost and accuracy of Proactive and Reactive Shepherding strategies with state-of-the-art LLM routing and cascading strategies. The routing baselines include RouteLLM~\citep{ong2024routellm} and GraphRouter~\citep{GraphRouter}, while the cascading baselines include FrugalGPT~\citep{chen2024frugalgpt} and ABC~\citep{kolawole2025agreementbased}. We use the vanilla Llama-3.2-3B-Instruct (without finetuning) as the SLM, deployed on a local server with an NVIDIA RTX 5090 GPU. For the LLM, we query Llama-3.3-70B-Versatile via the Groq API at \$0.59 per 1M input tokens and \$0.79 per 1M output tokens. The SLM is assigned a hardware-based cost of \$0.0058 per 1M tokens, derived from RTX 5090 throughput measurements and cloud rental rates (details in Appendix~\ref{app:slm_cost}).

\subsection{Datasets and Experimental Setup}

\textbf{Mathematical reasoning.}
GSM8K~\citep{cobbe2021gsm8k} contains 7,473 grade-school word problems for training and 1,319 for testing, each requiring multi-step arithmetic and paired with a numerical answer. We use 776 of the test questions for evaluation and the remaining 543 for validation. CNK12 is a Chinese K-12 mathematics benchmark drawn from OpenR1-Math-220k~\citep{openr1_math_220k}, covering algebra, geometry, and calculus from elementary through high school. We randomly sample 21,471 problems and split them into 18,251 for training, 1,073 for validation, and 2,147 for testing.

\textbf{Code generation.}
HumanEval~\citep{chen2021evaluatinglargelanguagemodels} has 164 programming problems, each with a function signature, docstring, and unit tests. MBPP~\citep{austin2021programsynthesislargelanguage} has 974 entry-level Python tasks, each with three test cases. Both are evaluated with the shepherding model trained on GSM8K, without any code-specific training, to test cross-domain transfer.

\textbf{Inference and Evaluation.}
Test sets contain no pre-computed hints, and hints are requested at inference time as described in Section~\ref{sec:training}. A response counts as correct when its final numerical answer matches the ground truth (GSM8K, CNK12) or when it passes all unit tests (HumanEval, MBPP). Each query is scored by majority vote over 7 sampled trials. Decoding settings, prompt templates, and scoring details are given in Appendix~\ref{app:inference-eval}. We compare methods from the widest view to the narrowest: full cost--accuracy curves (Section~\ref{sec:pareto}), the minimum cost at a fixed accuracy target (Section~\ref{min}), and a single default operating point per method (Section~\ref{sec:default-points}).

\subsection{Pareto Cost--Accuracy Analysis and AIQ}
\label{sec:pareto}

\begin{figure}[t]
    \centering
    \includegraphics[width=\linewidth]{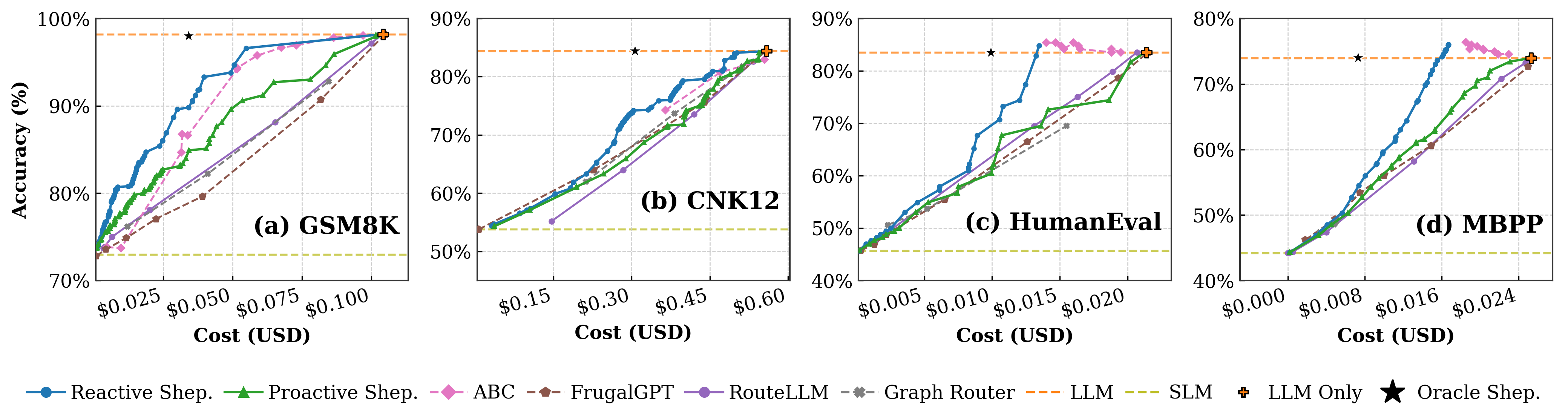}
    \caption{Cost--accuracy Pareto curves on the four benchmarks.}
    \label{fig:pareto}
    \vspace{-0.15in}
\end{figure}

Routing, cascading, and shepherding each expose knobs that trade cost against accuracy. We therefore sweep every method across its hyperparameter space and plot the resulting cost--accuracy curves on all four benchmarks (Figure~\ref{fig:pareto}). Reactive Shepherding lies on or above the baselines across most of the cost range, while Proactive Shepherding, which routes without SLM inference, remains above or on par with the routing baselines. On the coding benchmarks, both Reactive Shepherding and ABC exceed the LLM-only baseline due to the cascading mechanism analyzed in Appendix~\ref{app:pareto:exceed}.

\begin{table}[t]
\centering
\caption{AIQ (\%) of each method, computed from the operating points in Figure~\ref{fig:pareto}.}
\label{tab:aiq-comparison}
\footnotesize
\setlength{\tabcolsep}{4pt}
\begin{tabular}{@{}lcccccc@{}}
\toprule
& \multicolumn{3}{c}{\textbf{Routing}} & \multicolumn{3}{c}{\textbf{Cascading}} \\
\cmidrule(lr){2-4}\cmidrule(lr){5-7}
\textbf{Dataset} & RouteLLM & GraphRouter & Proactive Shep. & FrugalGPT & ABC & Reactive Shep. \\
\midrule
GSM8K     & 81.6 & 81.5 & 87.9 & 80.5 & 87.9 & \textbf{91.5} \\
CNK12     & 62.0 & 63.3 & 66.1 & 62.4 & 62.0 & \textbf{69.3} \\
HumanEval & 58.1 & 56.9 & 63.0 & 59.0 & 59.6 & \textbf{69.3} \\
MBPP      & 53.8 & 48.3 & 59.1 & 55.1 & 52.9 & \textbf{63.3} \\
\bottomrule
\end{tabular}
\vspace{-0.15in}
\end{table}

\nop{To summarize each curve in a single number, we report AIQ (Section~\ref{sec:problem_formulation}), which does not depend on any chosen operating point (its relation to ACE is discussed in Appendix~\ref{app:ace}).
The curves compare methods at every accuracy level. We next fix a single accuracy level and compare what each method must spend to reach it.}

Table~\ref{tab:aiq-comparison} summarizes each sweep by its AIQ (Eq.~\eqref{eq:AIQ}), the area under the budget envelope of its points in Figure~\ref{fig:pareto}. Reactive Shepherding attains the highest AIQ on all four benchmarks, and Proactive Shepherding the highest among routing-style methods. These comparisons span the whole accuracy range; we next fix a single accuracy level and compare what each method must spend to reach it.

\subsection{Minimum Accuracy Requirement}
\label{min}

\begin{figure}[t]
    \centering
    \includegraphics[width=\linewidth]{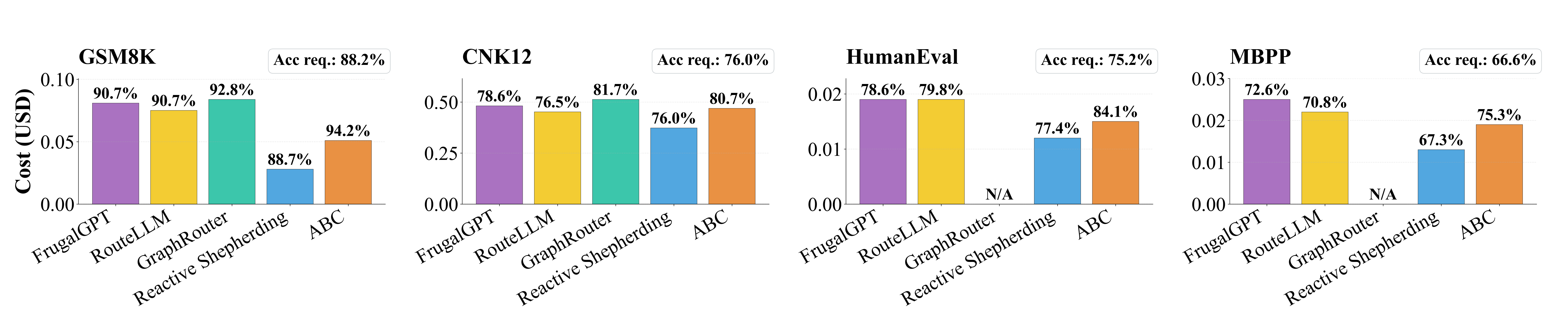}
    \caption{Minimum cost to achieve 90\% of LLM accuracy on each dataset. Each method is tuned to its optimal operating point, which meets or exceeds the accuracy requirement (shown above bars). N/A marks methods that do not reach the requirement at any configuration.}
    \label{fig:min-cost}
    \vspace{-0.15in}
\end{figure}

For each method, we solve Eq.~\eqref{eq:cost_optimal} with $\bar\tau$ set to 90\% of LLM accuracy on each dataset. Figure~\ref{fig:min-cost} compares minimum costs, and Appendix~\ref{app:method-configs} lists method configuration. Reactive Shepherding is cheapest across all benchmarks. On GSM8K (target 88.2\%), it reaches 88.7\% at \$0.028, 45\% less than ABC (\$0.051), which overshoots at 94.2\%. FrugalGPT reaches 90.7\% but costs \$0.082, nearly three times as much. On CNK12 (target 76.0\%), Reactive Shepherding costs \$0.37; ABC and FrugalGPT cost 27\% and 30\% more. On HumanEval (target 75.2\%), it costs \$0.012, 20\% below ABC and 37\% below RouteLLM. On MBPP, it costs \$0.013, 32\% below ABC (\$0.019).

\subsection{Comparison at Default Operating Points}
\label{sec:default-points}
Finally, we report every method at a single default operating point. Tables~\ref{tab:cost-analysis-gsm8k}--\ref{tab:cost-analysis-mbpp} summarize results across four benchmarks. For the baseline routing and cascading strategies, we use default parameter settings released by the authors in their respective GitHub repositories. For Proactive and Reactive Shepherding, we present the values for $\eta_\text{hint}$, $\alpha$, and $K$ in Appendix \ref{app:method-configs}. Reactive Shepherding achieves the highest ACE on all datasets, ranging from 1.09 (CNK12) to 1.89 (GSM8K), while reducing costs by 33--65\% relative to LLM-only inference. This ranking agrees with the AIQ and fixed-target comparisons above.
\begin{table}[t]
\centering
\setlength{\tabcolsep}{2pt}

\begin{minipage}[t]{0.48\linewidth}
    \centering
    \caption{Cost--accuracy on GSM8K. \\ LLM (\$0.104, 98.0\%) \textbar\ SLM (\$0.0008, 73.0\%).}
    \label{tab:cost-analysis-gsm8k}
    \resizebox{\linewidth}{!}{%
    \begin{tabular}{@{}lcccc@{}}
        \toprule
        \textbf{Strategy} & \textbf{Cost (\$)} & \textbf{Acc.\ (\%)} & \textbf{Red.\ (\%)} & \textbf{ACE} \\
        \midrule
        \multicolumn{5}{@{}l}{\textbf{Routing}} \\
        RouteLLM        & 0.064 & 88.1 & 38.5 & 0.99 \\
        GraphRouter     & 0.041 & 82.2 & 60.6 & 0.94 \\
        Proactive Shep. & 0.036 & 80.9 & 65.4 & 0.93 \\
        \midrule
        \multicolumn{5}{@{}l}{\textbf{Cascading}} \\
        FrugalGPT       & 0.039 & 79.6 & 62.5 & 0.71 \\
        ABC             & 0.051 & \textbf{94.3} & 51.0 & 1.75 \\
        Reactive Shep.  & \textbf{0.036} & 89.1 & \textbf{65.4} & \textbf{1.89} \\
        \midrule
        Oracle Shep.    & 0.034 & 98.0 & 67.3 & 3.11 \\
        \bottomrule
    \end{tabular}}
\end{minipage}
\hfill
\begin{minipage}[t]{0.48\linewidth}
    \centering
    \caption{Cost--accuracy on CNK12. \\ LLM (\$0.559, 84.4\%) \textbar\ SLM (\$0.006, 53.8\%).}
    \label{tab:cost-analysis-cnk12}
    \resizebox{\linewidth}{!}{%
    \begin{tabular}{@{}lcccc@{}}
        \toprule
        \textbf{Strategy} & \textbf{Cost (\$)} & \textbf{Acc.\ (\%)} & \textbf{Red.\ (\%)} & \textbf{ACE} \\
        \midrule
        \multicolumn{5}{@{}l}{\textbf{Routing}} \\
        RouteLLM        & 0.533 & \textbf{82.6} & 4.7 & 0.99 \\
        GraphRouter     & 0.513 & 81.7 & 8.2  & 0.99 \\
        Proactive Shep. & 0.489  & 80.6 & 12.5 & 1.00 \\
        \midrule
        \multicolumn{5}{@{}l}{\textbf{Cascading}} \\
        FrugalGPT       & 0.440 & 75.6 & 21.3 & 0.91 \\
        ABC             & 0.469 & 80.7 & 16.1 & 1.05 \\
        Reactive Shep.  & \textbf{0.373} & 76.0 & \textbf{33.3} & \textbf{1.09}   \\
        \midrule
        Oracle Shep.    & 0.309 & 84.4 & 44.7 & 1.83 \\
        \bottomrule
    \end{tabular}}
\end{minipage}
\vspace{-0.15in}
\end{table}

\begin{table}[t]
\centering
\setlength{\tabcolsep}{2pt}

\begin{minipage}[t]{0.48\linewidth}
    \centering
    \caption{Cost--accuracy on HumanEval.  \\LLM (\$0.0214, 83.5\%)\textbar\ SLM(\$0.00019, 45.7\%). }
    \label{tab:cost-analysis-humaneval}
    \resizebox{\linewidth}{!}{%
    \begin{tabular}{@{}lcccc@{}}
        \toprule
        \textbf{Strategy} & \textbf{Cost (\$)} & \textbf{Acc.\ (\%)} & \textbf{Red.\ (\%)} & \textbf{ACE} \\
        \midrule
        \multicolumn{5}{@{}l}{\textbf{Routing}} \\
        RouteLLM        & 0.020 & 83.5 & 6.5 & 1.07 \\
        GraphRouter     & 0.015 & 69.5 & 29.9 & 0.90 \\
        Proactive Shep. & \textbf{0.011} & 67.7 & \textbf{48.6} & 1.14 \\
        \midrule
        \multicolumn{5}{@{}l}{\textbf{Cascading}} \\
        FrugalGPT       & 0.019 & 78.66 & 11.2 & 0.98 \\
        ABC             & 0.015 & \textbf{84.15} & 29.9 & 1.46 \\
        Reactive Shep.  & 0.012 & 77.4 & 43.9 & \textbf{1.51} \\
        \midrule
        Oracle Shep.    & 0.009 & 83.5 & 57.9 & 2.41 \\
        \bottomrule
    \end{tabular}}
\end{minipage}
\hfill
\begin{minipage}[t]{0.48\linewidth}
    \centering
    \caption{Cost--accuracy on MBPP. \\ LLM (\$0.025, 74.0\%) \textbar\ SLM (\$0.00019, 44.2\%).}
    \label{tab:cost-analysis-mbpp}
    \resizebox{\linewidth}{!}{%
    \begin{tabular}{@{}lcccc@{}}
        \toprule
        \textbf{Strategy} & \textbf{Cost (\$)} & \textbf{Acc.\ (\%)} & \textbf{Red.\ (\%)} & \textbf{ACE} \\
        \midrule
        \multicolumn{5}{@{}l}{\textbf{Routing}} \\
        RouteLLM        & 0.013 & 58.2 & 48.0 & 0.91 \\
        GraphRouter     & \textbf{0.005} & 49.0 & \textbf{80.0} & 0.83 \\
        Proactive Shep. & 0.018 & 68.0 & 28.0 & 1.11 \\
        \midrule
        \multicolumn{5}{@{}l}{\textbf{Cascading}} \\
        FrugalGPT       & 0.015 & 60.6 & 40.0 & 0.92 \\
        ABC             & 0.019 & \textbf{75.3} & 24.0 & 1.38 \\
        Reactive Shep.  & 0.014 & 69.8 & 44.0 & \textbf{1.54} \\
        \midrule
        Oracle Shep.    & 0.007 & 74.0 & 72.0 & 3.64 \\
        \bottomrule
    \end{tabular}}
\end{minipage}
\vspace{-0.15in}
\end{table}

\textit{Comparison with routing.} 
Routing baselines (RouteLLM, GraphRouter) struggle when query difficulty varies widely. On CNK12, they achieve only 5--8\% cost reduction compared to LLM, whereas Reactive Shepherding delivers 33\%, four times the reduction, while still meeting the 90\% accuracy target. The gap is smaller on GSM8K, where most queries are SLM-solvable, but shepherding still nearly matches Oracle performance (\$0.036 vs.\ \$0.034) while RouteLLM incurs 78\% higher cost.

\textit{Comparison with cascading.}
Cascading baselines (FrugalGPT, ABC) achieve higher accuracy than routing but at a greater cost. On GSM8K, ABC reaches 94.3\% accuracy, 5.2 points above Reactive Shepherding---but costs 42\% more, yielding lower ACE (1.75 vs.\ 1.89). This illustrates a key trade-off: cascading pays for accuracy that may exceed application requirements, while shepherding optimizes cost-efficiency by requesting only partial LLM assistance.

\textit{Cross-domain generalization.}
HumanEval tests whether shepherding transfers to unseen domains: all methods use GSM8K-trained models without code-specific fine-tuning. Reactive Shepherding achieves 77.4\% accuracy while delivering 44\% cost reduction, yielding the highest ACE (1.51). 


\subsection{Latency Overhead: Proactive vs Reactive}

We measured inference latency on CNK12 using Llama-3.2-3B-Instruct on an RTX 5090 GPU. The shepherding policy adds only 7.32 ms per query, negligible versus SLM generation (384.62 ms). Proactive Shepherding makes hint decisions before SLM inference with minimal latency. Reactive Shepherding first generates $K=3$ SLM responses for consensus and features (1,163.48 ms), then runs the decision module (7.32 ms), a 0.6\% overhead. Thus, while Reactive Shepherding achieves better cost--accuracy trade-offs, Proactive Shepherding makes decisions $159\times$ faster locally.

\section{Conclusion} \label{sec:conclusion}

We have challenged the prevailing binary paradigm of routing and cascading, which treats LLMs as all-or-nothing resources. By introducing LLM Shepherding, we demonstrate that the most efficient path toward high-quality inference is not to choose between models, but to enable targeted collaboration. Our framework leverages token-level budget control, requesting only a short hint (prefix) from a frontier LLM to improve the accuracy of the SLM.

Our findings establish LLM Shepherding as a new approach for cost-efficient deployment. Across full cost--accuracy sweeps, Reactive Shepherding attains the highest AIQ on all four benchmarks, and at a fixed target of 90\% of LLM accuracy it is the cheapest method on all four. At default operating points it achieves the highest ACE on every benchmark and reduces cost by 33--65\% relative to LLM-only inference, up to four times the reduction of routing baselines and 1.5 times that of cascading baselines while meeting the 90\% accuracy target. The predictor also generalizes: trained solely on math data, it transfers zero-shot to code generation, achieving the highest ACE on HumanEval and MBPP, and it transfers to a second model pair without retraining (Appendix~\ref{app:cross-model-pipeline}).

Our evaluation has focused on verifiable reasoning tasks where correctness can be automatically assessed; extending shepherding to open-ended tasks such as summarization or creative writing would require replacing the binary correctness function with learned reward models or human evaluation, which we leave for future work.




\bibliography{refs}
\bibliographystyle{iclr2027_conference}

\newpage
\appendix

\section{Cross-Model Validation of the Hint-Size Effect}
\label{appendix:cross-model}

The motivating observation behind LLM Shepherding—that short LLM prefixes substantially improve SLM accuracy (Figure~\ref{fig:hints-help})—is established using a single Llama SLM-LLM pair. To check whether this effect is specific to that pair, we reproduce the hint-size vs.\ accuracy curve under three additional configurations on the mathematical reasoning benchmarks. These experiments validate the hint-size effect only, without training or evaluating the shepherding model. Appendix~\ref{app:cross-model-slm-size} additionally varies the SLM size, and Appendix~\ref{app:cross-model-pipeline} evaluates the full shepherding pipeline on a second model pair. The configurations are listed in Table~\ref{tab:cross-model-settings}.

\begin{table}[h]
\centering
\caption{Additional SLM-LLM configurations used to validate the hint-size effect. All models are used off-the-shelf without any alignment fine-tuning.}
\label{tab:cross-model-settings}
\small
\begin{tabular}{llll}
\toprule
Setting & SLM & LLM & Evaluation Data \\
\midrule
Unquantized Llama   & Llama-3.2-3B-Instruct & Llama-3.3-70B-Instruct$^\dagger$ & 2{,}000 CNK12 samples \\
Within-family Gemma & Gemma-3-4B-IT         & Gemma-3-27B-IT                   & GSM8K (full) \\
Cross-family        & Llama-3.2-3B-Instruct & Gemma-3-27B-IT                   & GSM8K (full) \\
Cross-family        & Llama-3.2-3B-Instruct & Gemma-3-27B-IT                   & 5{,}000 CNK12 samples \\
\bottomrule
\multicolumn{4}{l}{\footnotesize $^\dagger$ Default open-source weights served via Amazon Bedrock, without quantization.} \\
\end{tabular}
\end{table}

\begin{figure}[h]
\centering
\begin{subfigure}[t]{0.48\linewidth}
    \centering
    \includegraphics[width=\linewidth]{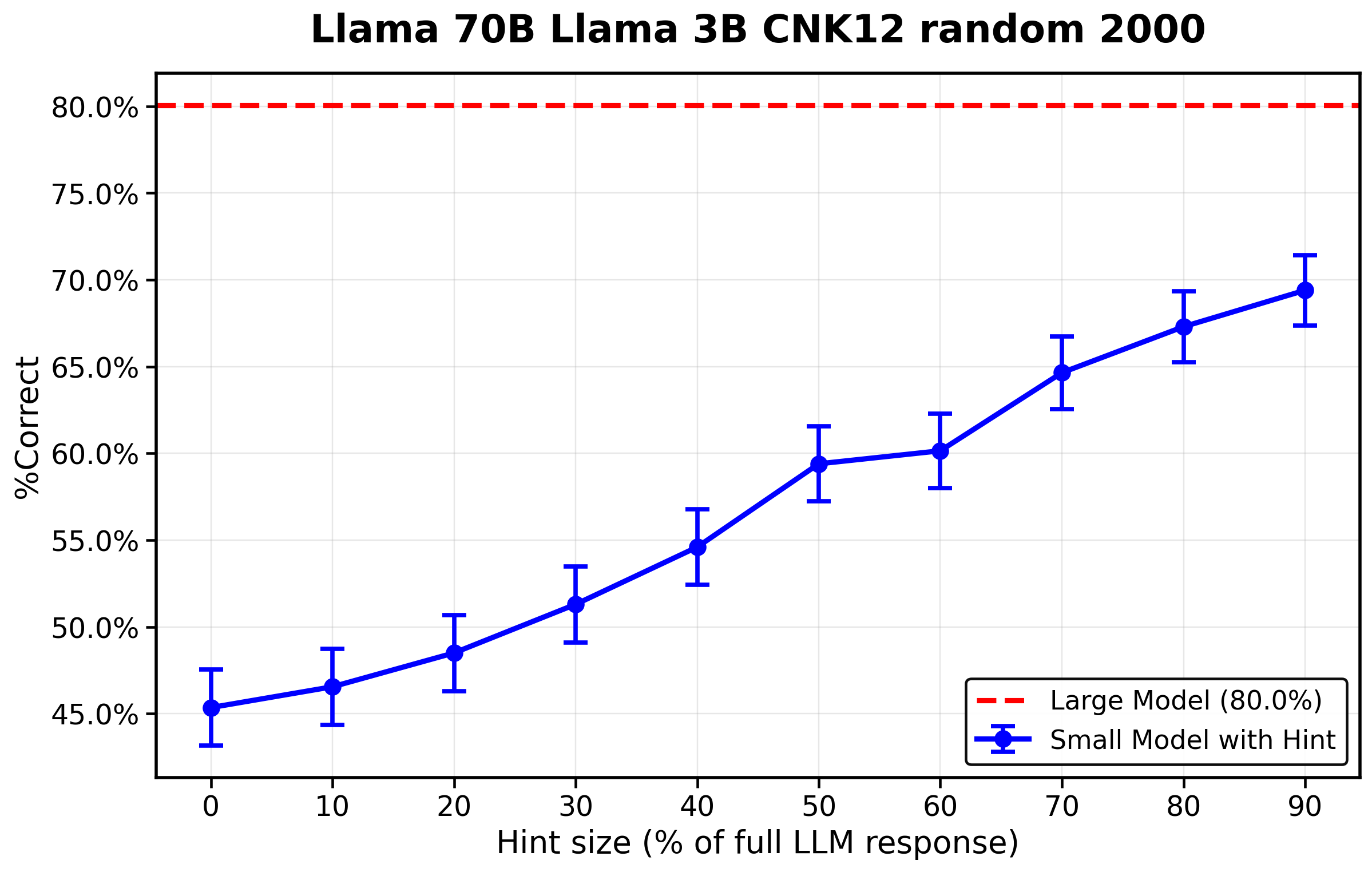}
    \caption{Unquantized Llama on CNK12.}
    \label{fig:cross-model-unquant}
\end{subfigure}
\hfill
\begin{subfigure}[t]{0.48\linewidth}
    \centering
    \includegraphics[width=\linewidth]{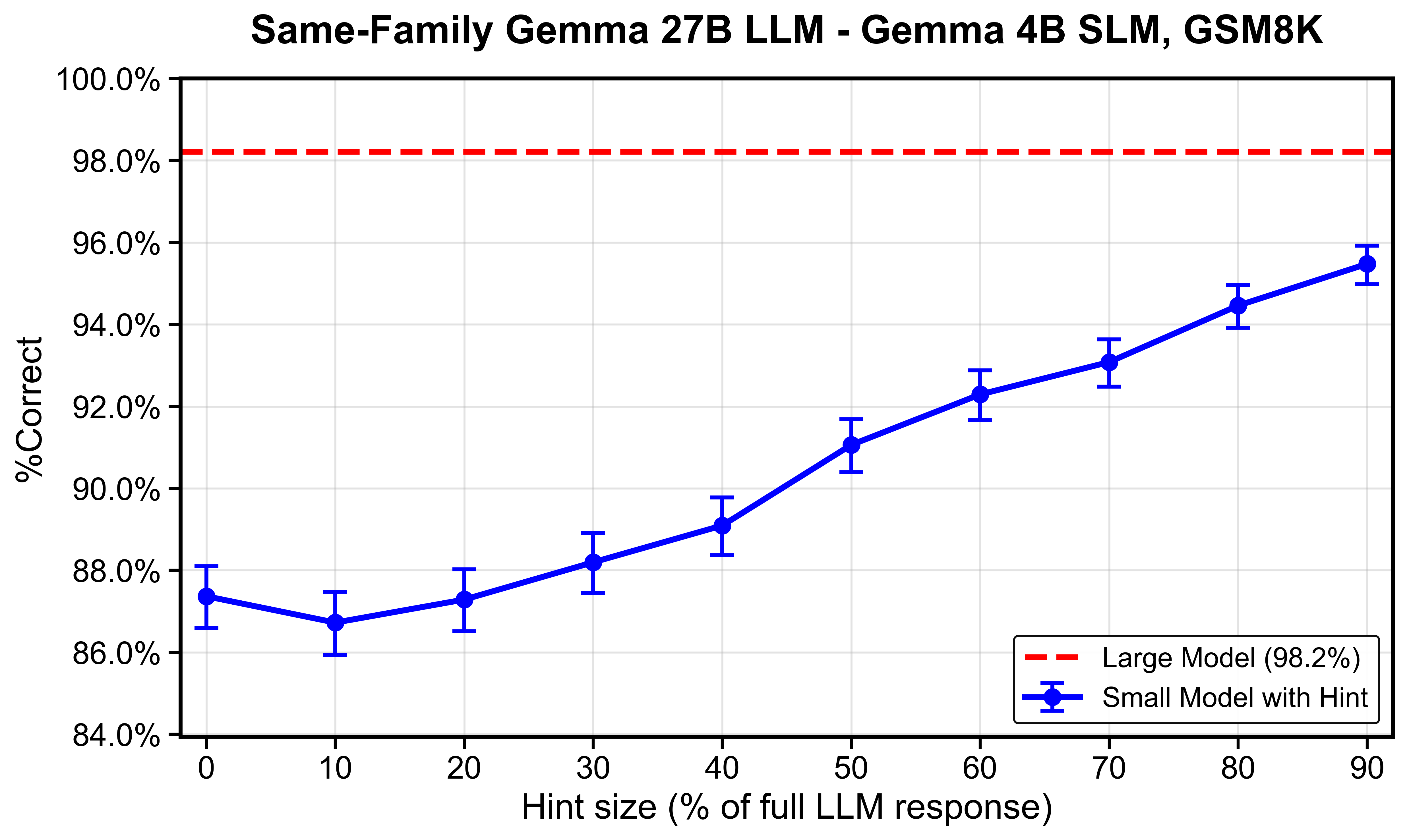}
    \caption{Within-family Gemma on GSM8K.}
    \label{fig:cross-model-gemma}
\end{subfigure}

\vspace{0.5em}

\begin{subfigure}[t]{0.48\linewidth}
    \centering
    \includegraphics[width=\linewidth]{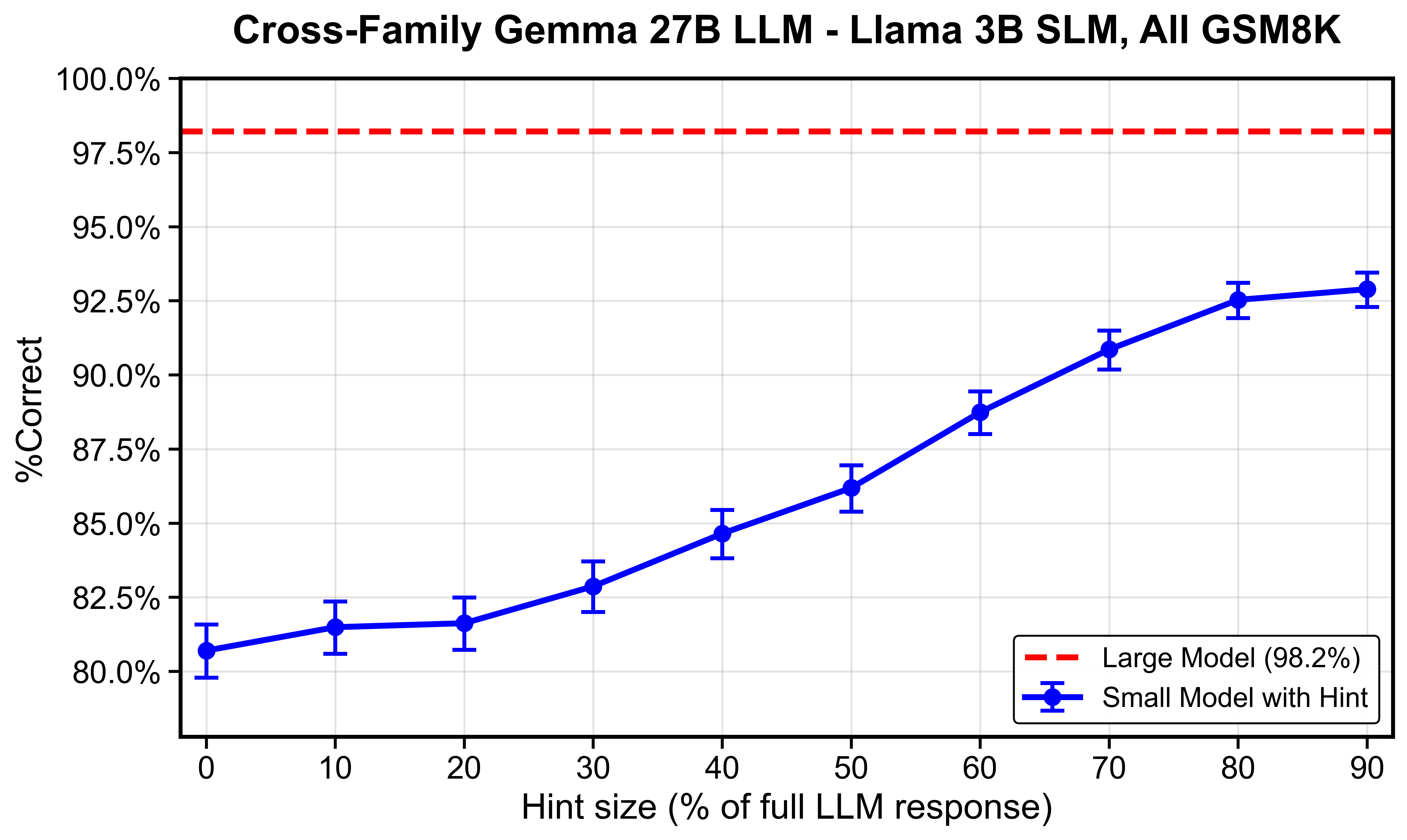}
    \caption{Cross-family on GSM8K.}
    \label{fig:cross-model-cross-gsm8k}
\end{subfigure}
\hfill
\begin{subfigure}[t]{0.48\linewidth}
    \centering
    \includegraphics[width=\linewidth]{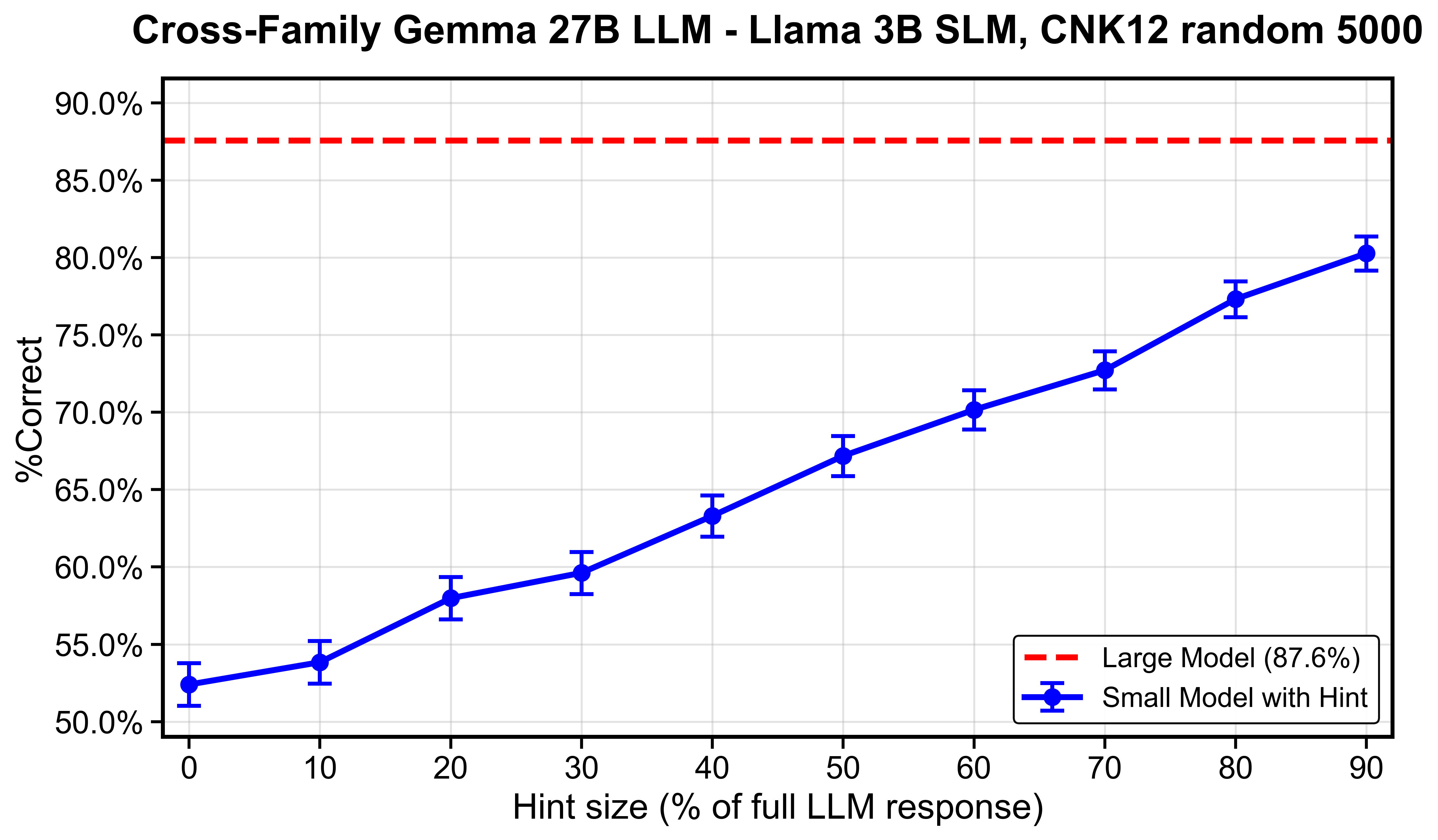}
    \caption{Cross-family on CNK12.}
    \label{fig:cross-model-cross-cnk12}
\end{subfigure}
\caption{SLM accuracy as a function of hint size across alternative SLM-LLM configurations. The broadly increasing trend from Figure~\ref{fig:hints-help} persists across quantization, model family, and tokenizer changes.}
\label{fig:cross-model-results}
\end{figure}

The \emph{Unquantized Llama} setting replaces the Groq-served quantized LLM with the official open-source weights, isolating the effect of provider-side quantization. The \emph{Within-family Gemma} setting swaps both models for a different family with a different tokenizer. The \emph{Cross-family} setting pairs a Llama SLM with a Gemma LLM, so the hint prefix and the SLM use different tokenizers and were trained independently. For each configuration, we sweep the hint size from 0\% to 90\% of the full LLM response in 10\% increments and measure SLM accuracy. Results are shown in Figure~\ref{fig:cross-model-results}.

In all four configurations, SLM accuracy increases approximately monotonically with hint size, matching the pattern in Figure~\ref{fig:hints-help}. We therefore conjecture that the hint-size effect is a general property of prefix-guided SLM generation, largely invariant to model family, tokenizer, and quantization, as long as both models are reasonably capable on the target task. This suggests LLM Shepherding can in principle be applied to heterogeneous SLM-LLM pairs. Appendix~\ref{app:cross-model-pipeline} evaluates the full pipeline on the within-family Gemma pair, and we leave the full pipeline evaluation under cross-family pairs to future work.

\subsection{Varying the SLM Size}
\label{app:cross-model-slm-size}

The hint-size effect may also depend on the capability of the SLM itself. We repeated the hint-size sweep on the GSM8K test set (776 queries) with three SLMs of increasing size. All three read the same Llama-3.3-70B hints. Table~\ref{tab:slm-size} reports the results.

\begin{table}[h]
\centering
\caption{SLM accuracy (\%) as a function of hint size for three SLM sizes on GSM8K. All SLMs read the same Llama-3.3-70B hints.}
\label{tab:slm-size}
\small
\begin{tabular}{lccc}
\toprule
Hint size & Llama-3.2-1B & Llama-3.2-3B & Llama-3.1-8B \\
\midrule
0\% (no hint) & 42.8 & 77.5 & 84.8 \\
10\% & 44.2 & 81.1 & 87.8 \\
20\% & 49.5 & 81.2 & 86.2 \\
30\% & 50.9 & 81.7 & 89.2 \\
40\% & 52.1 & 84.8 & 89.8 \\
50\% & 57.7 & 85.4 & 91.2 \\
60\% & 61.9 & 86.6 & 93.3 \\
70\% & 66.8 & 89.8 & 94.2 \\
80\% & 74.0 & 93.0 & 94.8 \\
90\% & 76.0 & 94.1 & 95.2 \\
\midrule
\textbf{Gain (0$\to$90\%)} & \textbf{+33.2} & \textbf{+16.6} & \textbf{+10.4} \\
\bottomrule
\end{tabular}
\end{table}

Hints help at all three scales. The absolute gain shrinks as the SLM improves. Relative to each model's headroom to the LLM's 98.0\%, however, the ordering changes: the 1B recovers about 60\% of its gap, while the 3B and 8B recover 81\% and 79\%. The weakest model needs hints most but exploits them least, and even a 90\% hint leaves it below what the 3B reaches unaided. At the other end, the 8B passes 89\% at a 30\% budget where the 3B needs 70\%: a stronger SLM needs a shorter hint for the same accuracy. Shepherding therefore does not depend on a narrow capability window.

\subsection{Full-Pipeline Validation on a Second Model Pair}
\label{app:cross-model-pipeline}

We ran the full pipeline on a second model pair: Gemma-3-4B-IT as the SLM and Gemma-3-27B-IT as the LLM. We reused the GSM8K-trained predictor from the Llama pair, with no retraining.

Table~\ref{tab:gemma-pipeline} follows the protocol of Section~\ref{min}: each method is reported at its cheapest configuration that reaches 90\% of the LLM's accuracy. This target follows RouteLLM~\citep{ong2024routellm}, which frames the routing objective as minimizing cost while reaching a set fraction of the stronger model's performance. Gemma-3-27B-IT scores 96.0\% on the 776-query GSM8K test set, the same split used for every method in this table, so the target accuracy is 86.4\%. The table groups methods by family: routing-style with Proactive Shepherding, then cascading-style with Reactive Shepherding. SLM-only and LLM-only are reference points.

\begin{table}[h]
\centering
\caption{Minimum cost to reach 90\% of the LLM's accuracy on GSM8K with the Gemma-3-4B-IT/Gemma-3-27B-IT pair, using the GSM8K-trained predictor from the Llama pair without retraining.}
\label{tab:gemma-pipeline}
\small
\begin{tabular}{lcc}
\toprule
Method & Cost (\$) & Accuracy \\
\midrule
SLM-only & 0.0012 & 86.21\% \\
RouteLLM & 0.0104 & 87.37\% \\
GraphRouter & 0.0226 & 88.79\% \\
\textbf{Proactive Shepherding} & 0.0280 & 88.02\% \\
\midrule
ABC & 0.0102 & 90.04\% \\
FrugalGPT & 0.0229 & 88.14\% \\
\textbf{Reactive Shepherding} & \textbf{0.0058} & 89.50\% \\
\midrule
LLM-only & 0.0563 & 96.0\% \\
\bottomrule
\end{tabular}
\end{table}

Every method reaches the target, and Reactive Shepherding does so at the lowest cost, 43\% below the next cheapest (ABC). Proactive Shepherding also reaches the target. Proactive mode acts on the predictor's output alone, with no consensus check to correct it, so meeting the target on an unseen model family shows that the hint-size decisions themselves transfer. Tables~\ref{tab:cost-analysis-humaneval}--\ref{tab:cost-analysis-mbpp} apply the same GSM8K-trained predictor to a different task domain. Together, these results suggest the predictor captures query difficulty rather than properties of one model pair or one dataset.
\section{Relation of ACE to Existing Cost-Efficiency Metrics}\label{app:ace}

ACE (Section~\ref{sec:problem_formulation}) summarizes each method at one operating point in the cost--accuracy plane. This appendix relates ACE to the metrics used in prior work and reports a complementary metric that does not depend on the choice of operating point. We do not claim ACE as a contribution of this work.

\paragraph{Relation to IBC.}
ACE coincides with the IBC metric of AutoMix~\citep{aggarwal2024automix} up to a fixed normalization. Writing both definitions out,
\[
\text{ACE}(\pi) = \frac{(A_{\pi} - A_s)/(A_l - A_s)}{(C_{\pi} - C_s)/(C_l - C_s)} = \frac{\mathrm{IBC}_{\pi}}{\mathrm{IBC}_{\mathrm{BASE}}},
\qquad
\mathrm{IBC}_{\mathrm{BASE}} = \frac{A_l - A_s}{C_l - C_s},
\]
so the two metrics rank methods identically. IBC therefore cannot serve as an independent check on ACE, and it grows without bound near the SLM cost exactly as ACE does.

\paragraph{Relation to PGR and APGR.}
The RouteLLM metrics~\citep{ong2024routellm} are related but measure different things. PGR is the fraction of the weak-to-strong accuracy gap that a policy recovers, which is exactly the accuracy part (numerator) of ACE. APGR averages PGR as the routing threshold is swept, and its cost axis is the fraction of queries sent to the strong model. That axis does not fit our problem setup: a 25-token hint and a 124-token full response both count as one strong-model call, so the token-level saving that shepherding targets is invisible there.

\paragraph{Behavior near the SLM cost.}
ACE grows without bound as a method's cost approaches the SLM cost $C_s$, where its denominator vanishes. We never tune our methods to maximize ACE, and every operating point reported in Tables~\ref{tab:cost-analysis-gsm8k}--\ref{tab:cost-analysis-mbpp} sits far from that boundary: Reactive Shepherding costs 45$\times$ the SLM cost on GSM8K, and 62$\times$, 63$\times$, and 74$\times$ on CNK12, HumanEval, and MBPP. ACE is also normalized within each dataset, so we never compare ACE values across datasets.

\paragraph{AIQ as an operating-point-free check.}
As a check that does not depend on any chosen operating point, we compute AIQ~\citep{hu2024routerbench}, the normalized area under the cost--accuracy curve, over the Pareto sweeps of Appendix~\ref{app:pareto}, which cover every method across its full hyperparameter range. The oracle policy is an upper bound rather than a runnable method, so we omit it. 

Reactive Shepherding has the highest AIQ on all four datasets, in agreement with the ACE comparison at the default operating points and with the Pareto curves of Appendix~\ref{app:pareto}. The three views are complementary: ACE summarizes the default operating points reported in the main tables, the Pareto curves compare methods at every quality level, and AIQ aggregates each curve into a single operating-point-free number.

\section{Proof of Corollary \ref{cor:dominance}}\label{proof:cor:dominance}
We treat the two cases $I_s(q) = 1$ and $I_s(q) = 0$ separately.

\paragraph{Case 1: $I_s(q) = 1$.}
The SLM alone produces a satisfactory response, so $n^*(q) = 0$. Substituting into Proposition~\ref{prop:oracle_costs} and using $\ind{n^*(q) > 0} = 0$ and $h_s^{(0)}(q) = h_s(q)$:
\begin{align*}
c_{\text{route}}^*(q) &= |q|\,c_s^{\text{in}} + |h_s(q)|\,c_s^{\text{out}}, \\
c_{\text{casc}}^*(q) &= |q|\,c_s^{\text{in}} + |h_s(q)|\,c_s^{\text{out}}, \\
c_{\text{proac-shep}}^*(q) &= |q|\,c_s^{\text{in}} + |h_s(q)|\,c_s^{\text{out}}, \\
c_{\text{reac-shep}}^*(q) &= |q|\,c_s^{\text{in}} + |h_s(q)|\,c_s^{\text{out}}.
\end{align*}
All four costs coincide, and both inequalities hold with equality.

\paragraph{Case 2: $I_s(q) = 0$.}
The SLM alone fails, so $n^*(q) > 0$ and the indicator $\ind{n^*(q) > 0} = 1$. Substituting into Proposition~\ref{prop:oracle_costs}:
\begin{align*}
c_{\text{route}}^*(q) &= |q|\,c_l^{\text{in}} + |h_l(q)|\,c_l^{\text{out}}, \\
c_{\text{casc}}^*(q) &= |q|\,c_s^{\text{in}} + |h_s(q)|\,c_s^{\text{out}} + |q|\,c_l^{\text{in}} + |h_l(q)|\,c_l^{\text{out}}, \\
c_{\text{proac-shep}}^*(q) &= |q|\,c_l^{\text{in}} + n^*(q)\,c_l^{\text{out}} + (|q| + n^*(q))\,c_s^{\text{in}} + |h_s^{(n^*(q))}(q)|\,c_s^{\text{out}}, \\
c_{\text{reac-shep}}^*(q) &= |q|\,c_s^{\text{in}} + |h_s(q)|\,c_s^{\text{out}} + |q|\,c_l^{\text{in}} + n^*(q)\,c_l^{\text{out}} \\
&\quad + (|q| + n^*(q))\,c_s^{\text{in}} + |h_s^{(n^*(q))}(q)|\,c_s^{\text{out}}.
\end{align*}

\textit{Proactive vs.\ Routing.} The term $|q|\,c_l^{\text{in}}$ appears in both expressions. Subtracting it,
\[
c_{\text{proac-shep}}^*(q) \leq c_{\text{route}}^*(q)
\;\iff\;
n^*(q)\,c_l^{\text{out}} + (|q| + n^*(q))\,c_s^{\text{in}} + |h_s^{(n^*(q))}(q)|\,c_s^{\text{out}} \leq |h_l(q)|\,c_l^{\text{out}}.
\]
Rearranging gives exactly the dominance condition~\eqref{eq:dominance_condition}.

\textit{Reactive vs.\ Cascading.} The term $|q|\,c_s^{\text{in}} + |h_s(q)|\,c_s^{\text{out}} + |q|\,c_l^{\text{in}}$ appears identically in both expressions. Cancelling it leaves the same inequality as in the previous case, which is again equivalent to~\eqref{eq:dominance_condition}.

Therefore, both $c_{\text{proac-shep}}^*(q) \leq c_{\text{route}}^*(q)$ and $c_{\text{reac-shep}}^*(q) \leq c_{\text{casc}}^*(q)$ hold if and only if~\eqref{eq:dominance_condition} is satisfied.

\section{Training Shepherding System: Additional Details}

\subsection{Challenges in Predicting \texorpdfstring{$n^*(q)$}{n*(q)}}
\label{app:prediction_challenges}

Predicting $n^*(q)$ directly as an exact token count is challenging for several reasons.

\paragraph{Heavy-tailed hint sizes.} $n^*(q)$ varies significantly across queries---on GSM8K, 80.6\% of queries require no hint at all, while the remaining 19.4\% span a wide range from 10\% to 90\% of the full LLM response (see Appendix~\ref{app:heavy_tail} for a detailed discussion).

\paragraph{Non-smooth benefit.} A small change in $n$ can abruptly determine whether the SLM crosses the quality threshold $\tau$, as critical information (e.g., a key constraint or intermediate step) may appear at specific token boundaries.

\paragraph{Noisy accuracy curve.} Measured accuracy does not increase steadily with $n$ (cf.\ Fig.~\ref{fig:hints-help}). On HumanEval it dips at the 10\%, 30\%, and 60\% budgets, each within the error bars, and peaks at the largest budget (75.6\% at 90\%). The data show a non-smooth increase, not that large hints hurt, so the minimum sufficient size cannot be read off a smooth trend.

\paragraph{Tokenization sensitivity.} The same semantic content yields different token counts depending on the model and query phrasing, making direct regression to absolute token counts brittle.

These properties jointly motivate the two-step design (classify-then-regress) and the log-transformed regression target $r(q)=\log(1+n^*(q))$ used in Section~\ref{sec:training}.

\subsection{Training Dataset Construction Details}
\label{app:training_dataset}

This appendix provides the complete methodology for constructing the training dataset used to supervise the Shepherding model, including the discretized token budget approach and labeling procedure.

\subsubsection*{Supervision via Discretized Token Budgets}

We obtain supervision for the hint sizes by evaluating candidate hints at multiple granularities via token-budgeted LLM calls. For each training query $q$, we first query the LLM once to obtain a reference completion $h_l(q)$ and use its LLM-token length $|h_l(q)|$ to define percentage sizes $p\in\{0,10,20,\ldots,90\}$. We choose 10\% increments as they provide sufficient granularity to capture the non-smooth benefit curve of hint tokens while maintaining computational tractability—finer granularities (e.g., 5\% steps) would double the labeling cost with diminishing returns in identifying the minimal sufficient size, whereas coarser granularities (e.g., 25\% steps) risk missing critical transition points where hints become effective.

For each $p$, we compute the target token size:
\[
n_p=\left\lfloor \tfrac{p}{100}|h_l(q)|\right\rfloor
\]
and then re-query the LLM by setting the \texttt{max\_new\_tokens} parameter to $n_p$. This parameter acts as a hard constraint on the generation length, ensuring the model outputs a completion $h_l^{(n_p)}(q)$ that conforms to the pre-defined size $n_p$.

We treat this size-constrained completion as the candidate hint $\text{hint}(q,n_p)=h_l^{(n_p)}(q)$, and evaluate it by running the SLM on $q\oplus \text{hint}(q,n_p)$ and computing $\phi\!\bigl(q,h_s(q \oplus \text{hint}(q,n_p))\bigr)$.

We record the discretized minimal sufficient size
\[
n^*(q)=\min\left\{n_p:\phi\!\bigl(q,h_s(q \oplus \text{hint}(q,n_p))\bigr)\ge\tau\right\},
\]
which is a grid-based approximation to the continuous optimum. The case $p=0$ corresponds to $\text{hint}(q,0)=\emptyset$, i.e., the router chooses not to query the LLM for any hint and directly returns the SLM output $h_s(q)$. If none of the partial sizes up to $p=90$ meets the threshold, we set $n^*(q)=|h_l(q)|$ and use the full LLM response as the output.

\subsubsection*{Rationale for Design Choices}

\paragraph{10\% Granularity.} The choice of 10\% increments balances labeling efficiency with precision. Each training query requires $|\{0, 10, 20, \ldots, 90\}| = 10$ LLM calls to construct the supervision signal. Finer granularities would provide marginally better approximations to $n^*(q)$ but at high computational cost—5\% steps would require 20 calls per query, doubling the labeling budget. Conversely, coarser granularities (e.g., 25\% steps) risk missing critical transition points where the hint becomes sufficient, particularly for queries with sharp quality transitions.

\paragraph{Deterministic LLM Decoding.} For all hint-labeling LLM calls, we use deterministic decoding (\texttt{temperature}$=0$, \texttt{top\_p}$=1$) to ensure reproducible supervision signals and eliminate variance from sampling across different training runs. This design choice prioritizes consistency in the training labels over diversity in hint content.

\paragraph{Grid-based Approximation.} The discretized target $n^*(q)$ serves as a practical surrogate for a continuous optimum. While this introduces quantization error, the 10\% grid is sufficient to capture meaningful differences in hint requirements across queries, and the log-transformation of the regression target (Section~\ref{sec:training}) further smooths over small discretization artifacts.

\subsection{Model Architecture Details}
\label{app:architecture}

The numeric feature vector $\mathbf{f}(q)$ is passed through a lightweight MLP to produce $\mathbf{g}(q) = \mathrm{MLP}(\mathbf{f}(q))$, which is concatenated with the \texttt{[CLS]} embedding $\mathbf{h}_{\text{CLS}}(q)$ to form the fused representation $\mathbf{u}(q) = [\mathbf{h}_{\text{CLS}}(q); \mathbf{g}(q)]$. $\mathrm{Head}_{\text{hint}}$ and $\mathrm{Head}_{\text{size}}$ are small feed-forward networks attached to $\mathbf{u}(q)$: $\mathrm{Head}_{\text{hint}}$ maps $\mathbf{u}(q)$ to a single scalar logit for binary classification (hint-needed vs.\ no-hint), while $\mathrm{Head}_{\text{size}}$ maps $\mathbf{u}(q)$ to a single scalar for the regression target $r(q)$. To reduce variance in the hint-indicator logits, we apply multi-sample dropout and average logits across multiple passes \cite{inoue2019multisample}.

\subsection{Optimization and Implementation Details}
\label{opt}

We train with AdamW \cite{loshchilov2017decoupled} and maintain an exponential moving average (EMA) of parameters for evaluation and checkpoint selection \cite{moralesbrotons2024ema}.

\noindent\textbf{Class Balancing.} As noted in Section~\ref{sec:training}, we use weighted minibatch sampling to address class imbalance when the proportion of queries requiring hints deviates from 50\%. Concretely, we employ \texttt{WeightedRandomSampler} during training to balance the minibatch distribution, ensuring that both positive and negative examples are adequately represented in each gradient update and yielding robust decision boundaries.

\noindent\textbf{Policy Calibration.} Post-training, we perform a grid search on the validation set to calibrate the policy for a target cost size. Specifically, given a fixed cost constraint $C_{\text{target}}$, we search over the probability threshold $\alpha$ for the binary classification decision to identify the configuration that maximizes validation accuracy while satisfying the cost constraint. This allows us to adapt the trained model to different cost-accuracy operating points without retraining.

\subsection{Inference Routing Rationale}
\label{app:inference}

The hint-size threshold $\eta_{\text{hint}}$ separates low-complexity from high-complexity queries among those predicted to need a hint. Queries with $0 < \pi_\theta(q) \le \eta_{\text{hint}}$ are passed to the LLM with \texttt{max\_new\_tokens} $= \pi_\theta(q)$ to obtain $\text{hint}(q, \pi_\theta(q)) = h_l^{(\pi_\theta(q))}(q)$, and the shepherded response $h_s(q \oplus \text{hint}(q, \pi_\theta(q)))$ is returned. Queries with $\pi_\theta(q) > \eta_{\text{hint}}$ are deemed high-complexity and routed directly to the LLM, returning $h_l(q)$ without invoking the SLM.

This threshold-based routing captures the intuition that queries requiring extensive hints may be better served by the LLM's full reasoning capacity, avoiding the overhead of generating a lengthy hint only to invoke the SLM. The threshold $\eta_{\text{hint}}$ is a hyperparameter tuned on the validation set to balance cost and accuracy.

\subsection{Reactive Shepherding: Feature Engineering and Hyperparameter Choices}
\label{app:reactive_features}

This appendix supplements Section~\ref{sec:reactive} with the rationale behind the cascading hyperparameters and the precise definitions of the features fed into the shepherding model. Recall that for each query we run the SLM $K$ times with temperature sampling ($T=0.3$) to obtain candidate answers $\{A_1,\dots,A_K\}$ and predictive entropies $\{e_1,\dots,e_K\}$. We use stochastic sampling rather than deterministic decoding to obtain diversity in the responses, which lets us measure the SLM's epistemic uncertainty through both response variation and predictive entropy. Since we work with math and coding datasets, agreement is determined by exact match on final answers.

\paragraph{Choice of $K$ and $k$.} We set $K=3$ in our experiments as a practical balance between signal quality and computational cost. The choice of $k$ reflects a precision-cost trade-off: $k=3$ (unanimous agreement) is more conservative, reducing false negatives at the expense of higher routing rates and LLM costs, while $k=2$ (majority voting) is more aggressive, allowing more queries to bypass the LLM but risking more errors in direct SLM outputs. We explore both settings in our experiments (Section~\ref{sec:performance}).

\paragraph{Numeric Features.} Given the $K$ responses, we compute the following features and incorporate them into the shepherding model's numeric input $\mathbf{f}(q)$:
\begin{itemize}
    \item \textbf{Average entropy:} $\bar{e}(q) = \frac{1}{K}\sum_{i=1}^K e_i$, capturing the model's overall uncertainty;
    \item \textbf{Average output length:} $\bar{L}_{\text{out}}(q) = \frac{1}{K}\sum_{i=1}^K |A_i|$, reflecting the typical response complexity;
    \item \textbf{Query length:} $|q|$, measured in tokens via the backbone model's tokenizer.
\end{itemize}
We use token-level rather than character-level length features because tokenization aligns with the models' internal representations and provides a more semantically meaningful proxy for reasoning complexity; longer outputs typically indicate multi-step solutions requiring more intermediate reasoning. Token-based features also enable efficient batched processing without additional tokenization overhead at inference time. These features are standardized and fused with the text representation through a lightweight MLP.
\section{Outlier Analysis} \label{Outlier}
\subsection{Outlier Generation Strategy} 

To identify and filter outliers in our training datasets, we employ a consensus-based approach that leverages the agreement patterns across different hint sizes. For each sample, we generate SLM responses across 11 different hint-size configurations and compare them with the ground truth. At each hint size level, we classify the outcome as an outlier if the minority class (correct or incorrect) occurs, and as normal otherwise. We then aggregate these binary outlier flags to compute $k$, the total number of outliers per sample across all hint size settings.

\begin{figure}[htbp]
    \centering
    \includegraphics[width=0.7\linewidth]{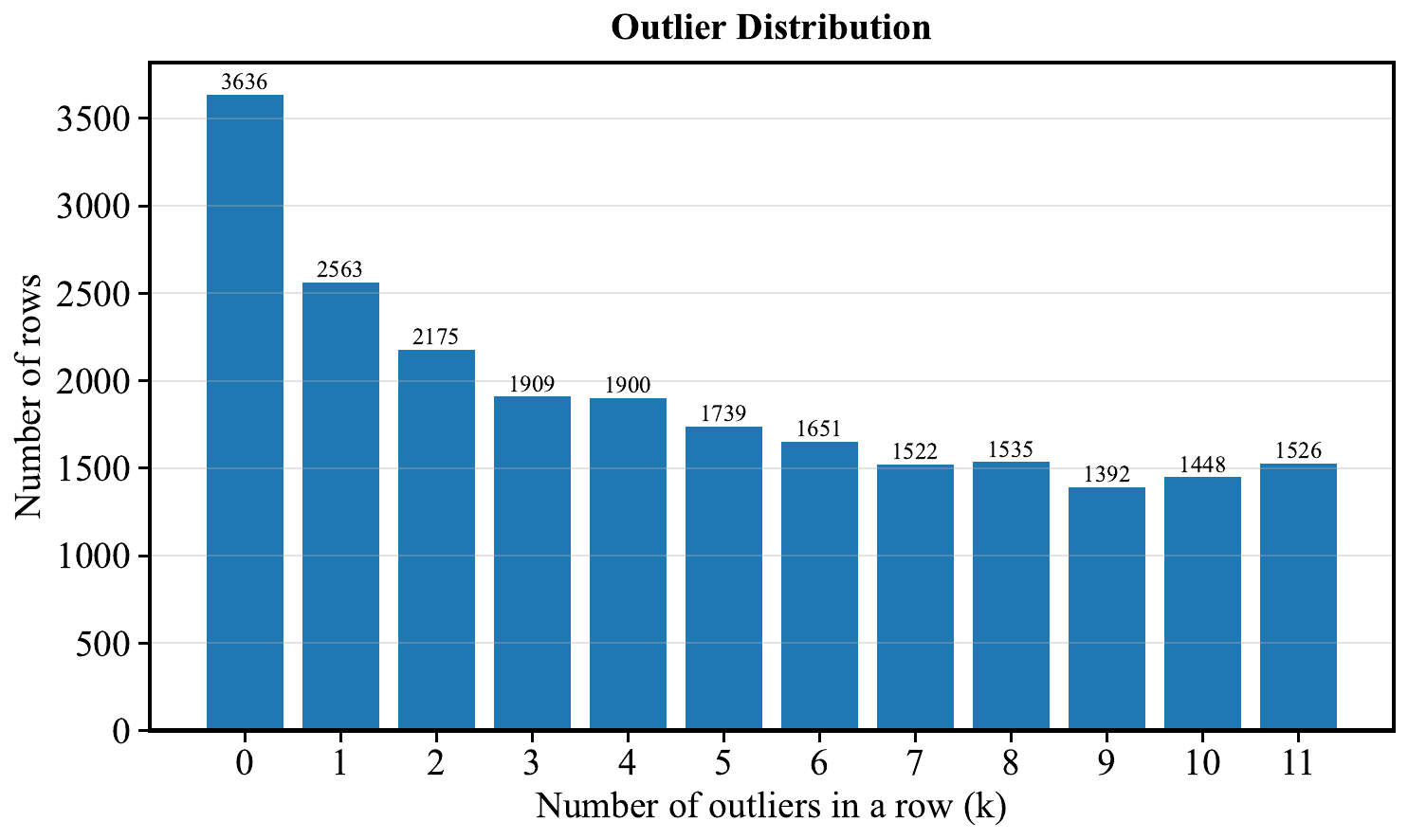}
    \caption{Outlier distribution for CNK12 dataset. }
    \label{fig:Outlier_CNK12}
\end{figure}

\begin{figure}[htbp]
    \centering
    \includegraphics[width=0.7\linewidth]{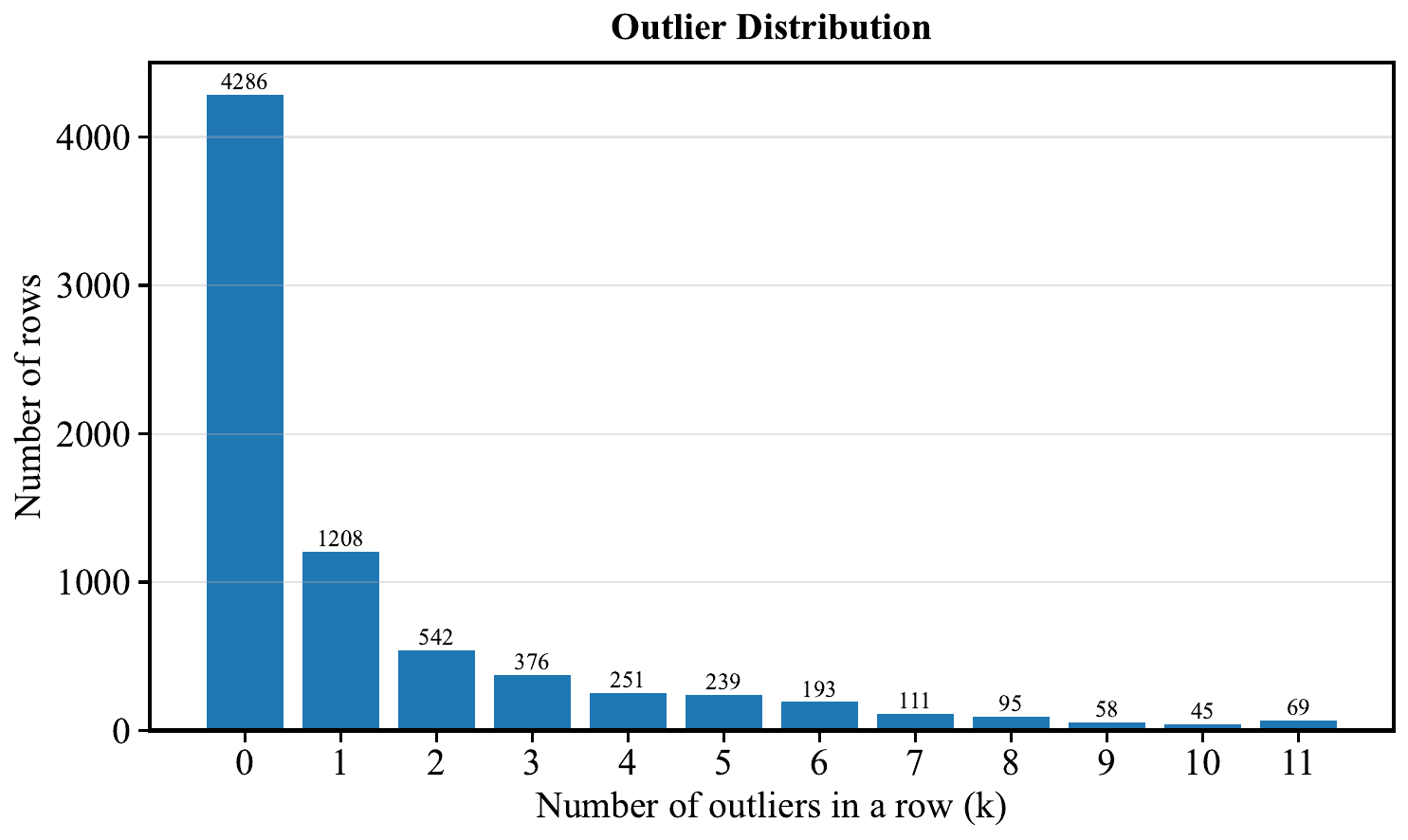}
    \caption{Outlier distribution for GSM8K dataset. }
    \label{fig:Outlier_GSM8K}
\end{figure}

Samples with high outlier counts (e.g., $k \geq \text{threshold}$) exhibit inconsistent behavior across varying hint sizes and may introduce noise during training. We therefore filter out such samples to refine the dataset quality. Figures~\ref{fig:Outlier_CNK12} and~\ref{fig:Outlier_GSM8K} illustrate the distribution of outlier counts for the CNK12 and GSM8K datasets, respectively.

Note that we apply this outlier filtering strategy only to CNK12 and GSM8K, as these datasets contain sufficient training samples to justify statistical filtering. For HumanEval, which consists of only 165 problems designed specifically for evaluation purposes, we do not perform outlier analysis.
    
\section{Zero-Inflated Distribution of Hint Sizes}
\label{app:heavy_tail}

\begin{figure}[t]
    \centering
    \includegraphics[width=0.7\linewidth]{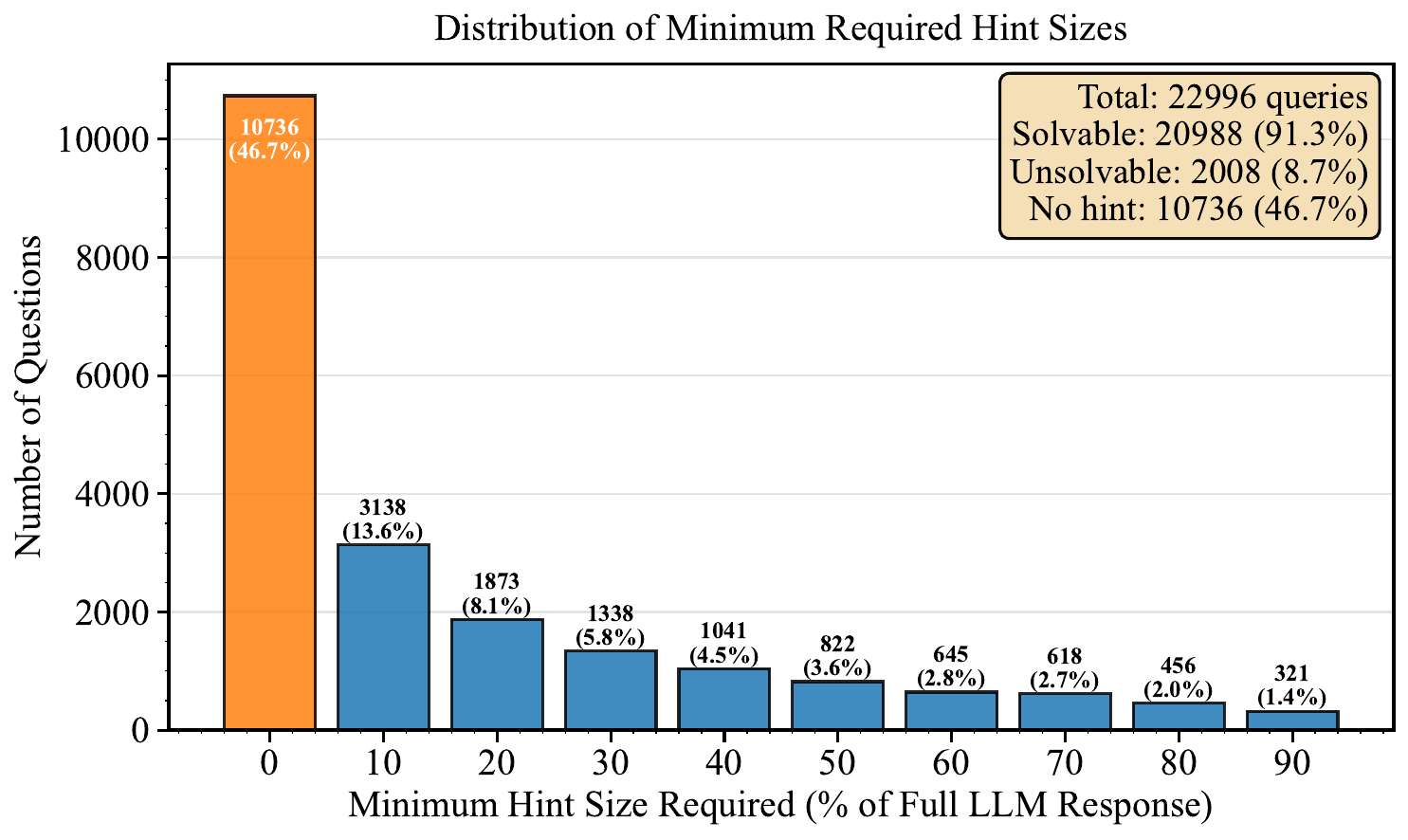}
    \caption{Distribution of minimum required hint sizes on CNK12 dataset (22,996 queries).}
    \label{fig:hint_distribution_cnk12}
\end{figure}

\begin{figure}[t]
    \centering
    \includegraphics[width=0.7\linewidth]{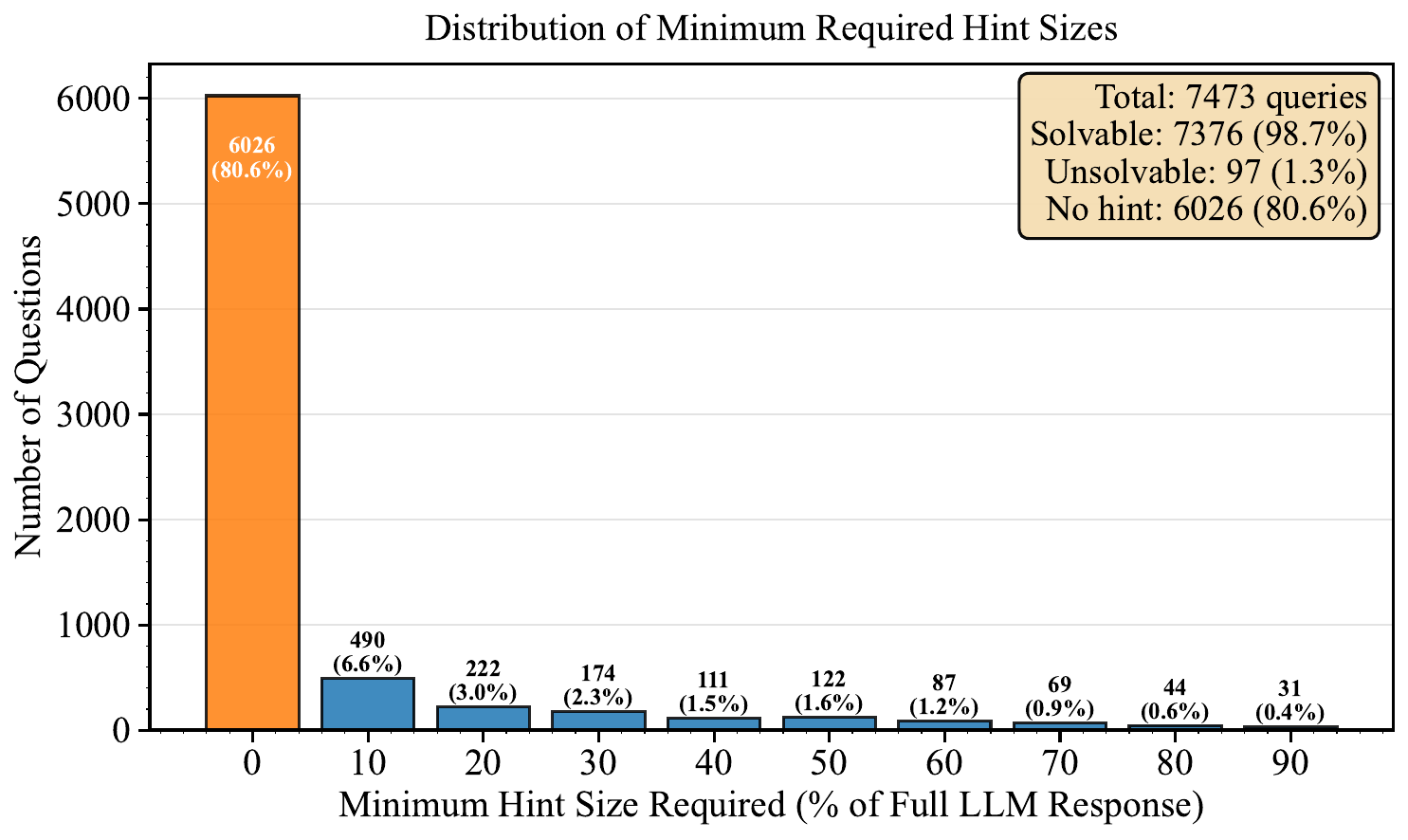}
    \caption{Distribution of minimum required hint sizes on GSM8K dataset (7,473 queries). }
    \label{fig:hint_distribution_gsm8k}
\end{figure}

As mentioned in Section~\ref{sec:training}, the distribution of required hint sizes is zero-inflated: a large point mass at $n^*(q) = 0$ together with a conditional distribution over positive hint sizes that decays approximately exponentially. Figure~\ref{fig:hint_distribution_cnk12} and Figure~\ref{fig:hint_distribution_gsm8k} illustrate this distribution across our training datasets CNK12 and GSM8K, respectively.

The distributions across both datasets demonstrate several key properties that motivate our two-stage prediction approach:

\begin{itemize}
    \item \textbf{Dataset-dependent class imbalance:} The proportion of queries requiring hints varies significantly between datasets. GSM8K shows extreme imbalance with 80.6\% (6,026 out of 7,473) requiring no hint, while CNK12 is more balanced with 46.7\% (10,736 out of 22,996) requiring no hint. This dataset-dependent behavior necessitates a flexible approach that can handle varying levels of class imbalance.
    
    \item \textbf{Consistent exponential decay in the positive tail:} Despite different absolute proportions, both datasets exhibit similar exponential decay patterns in their conditional hint size distributions. In CNK12, the distribution decays from 13.6\% (10\% hints) to 1.4\% (90\% hints), while in GSM8K it decays from 6.6\% (10\% hints) to 0.4\% (90\% hints). The consistent decay ratio (approximately 0.6--0.8$\times$ per step) confirms the exponential character of the conditional distribution.
    
    \item \textbf{High variance:} The standard deviation of hint sizes (computed only for queries requiring hints, i.e., $>0\%$) is substantial in both datasets. Directly regressing on raw token counts would be highly sensitive to these large values, motivating our log-transformation strategy: $r(q) = \log(1 + \tilde{n}^*(q))$ to compress the target range.
    
    \item \textbf{Variable unsolvability rates:} The proportion of queries unsolvable even with 90\% hints varies substantially: 8.7\% (2,008) in CNK12 versus only 1.3\% (97) in GSM8K. This indicates CNK12 contains more inherently challenging queries that exceed the SLM's capabilities even with substantial guidance. These queries are excluded from hint-size prediction because they require full LLM deployment.
    
    \item \textbf{Exponential decay characteristic:} Both datasets show clear exponential rather than linear or uniform decay. For example, in CNK12: 13.6\% → 8.1\% → 5.8\% → 4.5\%, and in GSM8K: 6.6\% → 3.0\% → 2.3\% → 1.5\%. This power-law-like behavior distinguishes the distribution from normal or uniform patterns and validates the need for specialized handling.
\end{itemize}

This zero-inflated distribution poses significant challenges for direct regression:
\begin{enumerate}
    \item A single regression model would struggle to accurately predict both the numerous no-hint cases (which constitute 47-81\% depending on the dataset) and the rare large-hint cases (which are critical for maintaining quality on difficult queries).
    \item Standard $\ell_2$ loss would be dominated by reconstruction error on the tail, potentially sacrificing accuracy on the more frequent short-hint cases.
    \item The varying class balance between datasets (47-53 split in CNK12 vs 81-19 split in GSM8K) requires flexible handling via weighted sampling to ensure robust decision boundaries across different data distributions.
    \item The presence of unsolvable queries (1.3-8.7\% depending on dataset) necessitates explicit handling to avoid the model attempting to predict hints for inherently infeasible cases.
\end{enumerate}

Our two-stage approach—first classifying whether any hint is needed, then predicting the size only for positive cases—naturally decomposes this distribution into two manageable sub-problems aligned with its zero-inflated structure: a balanced binary classification for the point mass at zero (via weighted sampling that adapts to dataset-specific imbalance) and a regression on the conditional distribution $p(n^* \mid n^* > 0)$, which excludes the dominant zero mass and focuses modeling capacity on the approximately exponential tail of positive hint sizes. This decomposition proves robust across datasets with varying difficulty profiles, as evidenced by the consistent performance on both the balanced CNK12 and the imbalanced GSM8K datasets.
\section{SLM Cost Calibration}
\label{app:slm_cost}

To make the cost comparison between locally hosted SLM inference and API-based LLM inference consistent, we assign the SLM a nonzero token-level cost based on measured hardware throughput. We benchmark \texttt{Llama-3.2-3B-Instruct} in BF16 using vLLM with continuous batching on the same RTX 5090 used in our experiments. We replay 7{,}473 evaluation queries and measure an end-to-end throughput of 17.7k tokens/s, corresponding to approximately 63.8M tokens/hour. Using an RTX 5090 cloud rental rate of \$0.37/hour (Vast.ai, May 2026), this yields an effective SLM cost of \$0.0058 per 1M tokens.

\paragraph{Sensitivity to the SLM price.}
The \$0.0058 per 1M token SLM price is an estimate, and it sets the bottom of the ACE ratio. To test how much the rankings depend on it, we priced the SLM at 1/100, 1/50, 1/20, and 1/10 of the LLM's input and output prices and recomputed both metrics on GSM8K, with everything else unchanged: ACE at the Table~\ref{tab:cost-analysis-gsm8k} operating points, and AIQ over the sweeps of Appendix~\ref{app:pareto}. The 1/100 point is close to the paper's estimate, since \$0.0058 is roughly 1/100 of the LLM's input price. The 1/10 point is close to Groq's listed price for the stronger Llama-3.1-8B (\$0.05 input and \$0.08 output per 1M tokens), so the band runs from our hardware-based estimate to what renting through the same API would cost.

\begin{table}[h]
\centering
\caption{AIQ (\%) on GSM8K as the SLM price varies from 1/100 to 1/10 of the LLM's input and output prices.}
\label{tab:slm-price-aiq}
\small
\begin{tabular}{lccc}
\toprule
SLM price ($\times$ LLM price) & Reactive Shepherding & ABC & FrugalGPT \\
\midrule
1/100 & 91.5 & 87.9 & 80.5 \\
1/50 & 91.5 & 87.6 & 80.4 \\
1/20 & 91.3 & 86.7 & 80.1 \\
1/10 & 91.1 & 85.0 & 79.6 \\
\bottomrule
\end{tabular}
\end{table}

\begin{table}[h]
\centering
\caption{ACE on GSM8K at the Table~\ref{tab:cost-analysis-gsm8k} operating points (cost and accuracy unchanged) as the SLM price varies from 1/100 to 1/10 of the LLM's input and output prices.}
\label{tab:slm-price-ace}
\small
\begin{tabular}{lcccccc}
\toprule
Strategy & Cost (\$) & Acc.\ (\%) & 1/100 & 1/50 & 1/20 & 1/10 \\
\midrule
FrugalGPT & 0.039 & 79.6 & 0.73 & 0.74 & 0.77 & 0.84 \\
ABC & 0.051 & 94.3 & 1.76 & 1.79 & 1.83 & 1.92 \\
Reactive Shep. & 0.036 & 89.1 & 1.91 & 1.94 & 2.06 & 2.23 \\
\bottomrule
\end{tabular}
\end{table}

The ordering between Reactive Shepherding and the cascading baselines does not change under either metric anywhere in this tenfold range. Most of the cost at these operating points comes from LLM output tokens, and all three methods pay the same SLM price, so raising it shifts them together instead of reordering them.
\section{Prompting and Evaluation Details} \label{prompt}

This section gives the prompt templates used for hint generation, SLM inference, and shepherding supervision, followed by the inference and evaluation protocol of Section~\ref{sec:performance}.

\subsection{Prompt Template}
Across all datasets, the small or large model receives the query with the same instructions and, when available, a generated large-model hint as additional contextual input. The general prompt structure is:
\begin{quote}
\small
\textbf{Instruction:} Produce a final answer following the task-specific constraints. \\
\textbf{Question:} \{query\} \\
\textbf{Hint:} \{generated large-model output\} \\
\end{quote}

When no hint is provided, the prompt is identical except that the \texttt{Hint:} line is omitted.

\textbf{GSM8K:} We use a fixed few-shot prompt consisting of two example question–answer pairs, followed by the target query. When available, a generated large-model hint is appended as additional context.

\textbf{MBPP:} The small and large models receive the function specifications, function signature, and provided test cases, along with the generated large-model hint if available. The model is instructed to generate only executable Python code, without additional explanation. The prompt structure otherwise follows the general template above.

\textbf{HumanEval:} Follows the same prompting procedure as MBPP; however, the test cases and function signature are already included in the dataset’s original prompt.

\textbf{CNK12:} Follows the same prompt template as GSM8K but uses single-shot prompting without example question–answer pairs.

\subsection{Inference and Evaluation Protocol}
\label{app:inference-eval}
Test sets contain no pre-computed hints. At inference time the policy predicts a hint size $\pi_\theta(q)$. When $\pi_\theta(q) > 0$, we query the LLM with \texttt{max\_new\_tokens} $= \pi_\theta(q)$ and append the returned prefix to the original prompt for SLM completion (Section~\ref{sec:training}). All test-time SLM and LLM generations use temperature $0.3$ and top-$p$ $0.95$.

For GSM8K and CNK12, we extract the final numerical answer from the response and compare it with the ground truth by exact match. For HumanEval and MBPP, we execute the generated program against the benchmark's unit tests and count a response as correct only when all tests pass. To reduce sampling variance, we repeat each query in 7 independent trials and report the majority-vote outcome.
\section{Method Configurations for Minimum Accuracy Experiments}
\label{app:method-configs}
Table~\ref{tab:method-configs} details the hyperparameter settings used for each method across all datasets in the minimum accuracy requirement experiments (Section~4.3). Each baseline is tuned to achieve accuracy closest to the 90\% LLM accuracy threshold on each dataset.
For RouteLLM, the values shown in the table are for the routing threshold. For FrugalGPT, the values represent the confidence threshold below which queries are escalated to the LLM. GraphRouter uses either its ``Performance First'' (P.F.) mode, which prioritizes accuracy, or an additional bias term added to increase cost orientation. ABC uses its default configuration with $N=2$, checking agreement between two SLM responses before escalating to the LLM. For Reactive Shepherding, $\eta_{\text{hint}}$ denotes the hint token budget; $K$ denotes the number of SLM responses used in consensus evaluation, and $\alpha$ denotes the probability threshold for the binary hint classification decision.
\begin{table*}[h]
\centering
\caption{Method configurations for minimum accuracy experiments. P.F.\ = Performance First mode; Bias = additive term to increase cost orientation.}
\label{tab:method-configs}
\setlength{\tabcolsep}{4pt}
\begin{tabular}{lcccc}
\toprule
\textbf{Method} & \textbf{GSM8K} & \textbf{CNK12} & \textbf{HumanEval} & \textbf{MBPP} \\
\midrule
RouteLLM & 0.48 & 0.57 & 0.70 & 0.58 \\
GraphRouter & P.F. & P.F. & P.F. + $4 \times 10^{-7}$ Bias & P.F. \\
FrugalGPT & 0.0006 & 0.001 & 0.001 & 0.00016 \\
ABC & $N=2$ & $N=2$ & $N=2$ & $N=2$ \\
Reactive Shepherding & \shortstack{$\eta_{\text{hint}} = 58$ \\ $K=3$, $\alpha=0.228$} & \shortstack{$\eta_{\text{hint}} = 60$ \\ $K = 2$, $\alpha=0.349$} & \shortstack{$\eta_{\text{hint}} = 110$ \\ $K=2$, $\alpha=0.228$} & \shortstack{$\eta_{\text{hint}} = 130$ \\ $K=2$, $\alpha=0.372$} \\
\bottomrule
\end{tabular}
\end{table*}
\section{Pareto Sweep Details}
\label{app:pareto}

Figure~\ref{fig:pareto} in Section~\ref{sec:pareto} plots the cost--accuracy curves of every method on all four benchmarks. This appendix details the sweep configurations behind those curves and analyzes the cascading methods that exceed the LLM-only baseline.

\subsection{Sweep Configurations}
\label{app:pareto:setup}

For each baseline and for Proactive and Reactive Shepherding, we vary the primary cost--accuracy trade-off parameter over a fine grid on the test set of each benchmark and record the resulting total cost and accuracy:
\begin{itemize}
    \item \textbf{RouteLLM}~\citep{ong2024routellm}: routing threshold swept over $[0,1]$.
    \item \textbf{GraphRouter}~\citep{GraphRouter}: cost-bias term swept, with Performance-First mode as the high-accuracy endpoint.
    \item \textbf{FrugalGPT}~\citep{chen2024frugalgpt}: cascading confidence threshold swept over its full operating range.
    \item \textbf{ABC}~\citep{kolawole2025agreementbased}: ensemble size $N$ and agreement requirement $k$ swept.
    \item \textbf{Proactive and Reactive Shepherding}: probability threshold $\alpha$ and hint budget $\eta_{\text{hint}}$ swept jointly; Reactive additionally varies the consensus requirement $k$ over $\{2,3\}$.
\end{itemize}

\subsection{Cascading Methods Exceeding the LLM-Only Baseline}
\label{app:pareto:exceed}

On HumanEval and MBPP, both Reactive Shepherding and ABC attain accuracy above the LLM-only baseline. This is initially counterintuitive, since the SLM is individually weaker than the LLM on both benchmarks. The behavior arises from the cascading mechanism. When the SLM achieves majority agreement across $K$ samples (with $K=2$ on both benchmarks in Table~\ref{tab:method-configs}), its consensus output is returned directly without invoking the LLM. On code generation tasks, this consensus output is occasionally correct on queries where the LLM's single-pass output fails to compile or contains subtle bugs. The system therefore combines two complementary sources of correct answers: SLM consensus on queries where agreement signals reliability, and LLM hints on harder queries. ABC, which uses SLM agreement as its gating mechanism, exhibits the same phenomenon for the same reason. The effect is specific to code generation, where stochastic single-shot LLM outputs can fail while stable SLM consensus succeeds; on GSM8K and CNK12, the LLM's single-pass accuracy is high enough that no method exceeds it.

\section{More Related Works}\label{sec:related}
Existing techniques for efficient, accurate, or low-cost SLM-LLM collaborative inference include: 1) LLM routing, 2) LLM cascading, 3) speculative decoding, and 4) knowledge distillation.
\subsection{LLM Routing}

LLM routing addresses cost-quality trade-offs by using a learned router to decide \emph{a priori} which model should handle a given query. Early work by \textbf{Lu et al.}~\cite{lu2023routing} introduced \textbf{Zooter}, a reward-guided routing method that distills rewards from training queries to learn a routing function, achieving top performance on 44\% of tasks across 26 benchmark subsets while maintaining computational efficiency. Building on this foundation, \textbf{Hybrid-LLM}~\cite{ding2024hybrid} proposed quality-aware query routing that dynamically trades quality for cost based on predicted query difficulty, achieving up to 40\% fewer calls to large models with no quality degradation. \textbf{RouterBench}~\cite{hu2024routerbench} subsequently established a standardized evaluation framework comprising over 405k inference outcomes from representative LLMs, enabling systematic comparison of routing strategies. \textbf{RouteLLM}~\cite{ong2024routellm} advanced the field by introducing a principled framework for learning routers from human preference data, demonstrating cost reductions of over 85\% on MT-Bench and 35\% on GSM8K while maintaining 95\% of GPT-4's performance. Most recently, \textbf{GraphRouter}~\cite{GraphRouter} proposed an inductive graph framework that models contextual interactions among tasks, queries, and LLMs as a heterogeneous graph, achieving at least 12.3\% improvement over prior routers and enhanced generalization to unseen LLMs.

Despite these advances, routing exhibits a fundamental structural limitation: the decision is inherently binary (SLM or LLM), and when the LLM is selected, the system incurs the full inference cost. Routing does not facilitate partial reuse of LLM knowledge, rendering it insufficiently flexible for strict cost constraints.

\subsection{LLM Cascading}


LLM cascading arranges models hierarchically, escalating queries to more capable models when weaker models fail quality criteria. \textbf{FrugalGPT}~\cite{chen2024frugalgpt} pioneered this approach, demonstrating up to 98\% cost reduction by learning which LLM combinations to use per query. \textbf{LLM-Blender}~\cite{jiang2023llm} introduced an ensemble perspective, using pairwise ranking to select among candidate outputs and generative fusion to merge top candidates. \textbf{AutoMix}~\cite{aggarwal2024automix} advanced cascading through few-shot self-verification and POMDP-based routing, reducing costs by over 50\%. Recent work has explored both learned and training-free approaches: \textbf{C3PO}~\cite{chen2025c3pocompactplugandplayproxy} provides label-free cascading with formal cost guarantees, while \textbf{Agreement-Based Cascading (ABC)}~\cite{kolawole2025agreementbased} proposes a simple, training-free strategy where escalation decisions are based on agreement among an ensemble of model responses. In our single-SLM setting, we adapt ABC by measuring consensus across multiple responses from the same SLM.

While cascading reduces average cost, it remains ill-suited for strict budget constraints: SLMs frequently fail on complex tasks, triggering full LLM inference. Each escalation requires a complete LLM forward pass, limiting cost savings when SLM reliability is low.

\subsection{Speculative Decoding}

Speculative decoding~\cite{leviathan2023fast,chen2023acceleratinglargelanguagemodel} is a widely adopted technique for accelerating LLM inference without sacrificing output quality. The approach uses a smaller \emph{draft model} to generate candidate token sequences, which are then verified in parallel by the larger \emph{target model}. Tokens that match the target model's distribution are accepted, amortizing the cost of large-model inference across multiple tokens. \textbf{Medusa}~\cite{cai2024medusa} extends this idea by adding multiple decoding heads to the LLM itself, eliminating the need for a separate draft model. \textbf{Draft \& Verify}~\cite{zhang-etal-2024-draft} achieves self-speculative decoding by selectively skipping intermediate layers during drafting, providing up to 2$\times$ speedup without additional training. \textbf{Online Speculative Decoding}~\cite{Xiaoxuan2024} continuously adapts the draft model during inference to improve token acceptance rates.

Speculative decoding and shepherding solve fundamentally different problems. Speculative decoding accelerates LLM inference \emph{latency} by parallelizing token verification, but the LLM remains in the critical path for every token---the total LLM computation is unchanged. Shepherding reduces LLM \emph{monetary cost} by minimizing the number of LLM tokens generated. The two are complementary: shepherding could use speculative decoding to accelerate the hint-generation step itself. A direct empirical comparison would require equating their different objectives (latency vs.\ cost), which we leave as future work.

\subsection{Knowledge Distillation}

Knowledge distillation (KD)~\cite{hinton2015distilling} transfers capabilities from a large teacher model to a smaller student model during training. Recent work has extended KD to LLMs: \textbf{MiniLLM}~\cite{gu2024minillm} proposes reverse KL divergence as the distillation objective, which better suits generative language models by preventing the student from overestimating low-probability regions. \textbf{Distilling Step-by-Step}~\cite{hsieh2023distillingstepbystepoutperforminglarger} extracts chain-of-thought rationales from LLMs to train smaller models more data-efficiently, enabling a 770M-parameter model to outperform few-shot prompted 540B PaLM. Comprehensive surveys~\cite{gou2021knowledge,xu2024survey} provide extensive coverage of KD techniques for LLMs, including skill-specific and domain-specific distillation.

Knowledge distillation and shepherding are complementary but distinct. KD is a \emph{training-time} technique: once distillation is complete, the student model operates independently at inference. Shepherding, in contrast, is an \emph{inference-time} collaboration that dynamically injects LLM hints based on each query's difficulty. KD improves the SLM's baseline capability, while shepherding provides targeted assistance when that baseline proves insufficient.

\subsection{Positioning of LLM Shepherding}

Our work differs fundamentally from all the above approaches. Unlike routing and cascading, which make binary decisions about \emph{which} model generates the complete response, shepherding controls \emph{how much} LLM output to request, enabling a continuous trade-off between cost and quality. Unlike speculative decoding, which accelerates LLM inference while keeping the LLM in the critical path for every token, shepherding removes the LLM from final response generation entirely. Unlike knowledge distillation, which operates during training, shepherding provides dynamic, query-specific guidance at inference time. 

Shepherding complements existing approaches: proactive shepherding enhances routing by replacing full LLM calls with partial hints, reactive shepherding enhances cascading by substituting full escalations with targeted guidance, and shepherding can be combined with distilled SLMs for further improvements. To our knowledge, this is the first work to exploit token-level budget control for SLM-LLM collaboration.

\end{document}